\documentclass[12pt]{article}

\usepackage{scicite}

\usepackage{times}
\usepackage[utf8]{inputenc}
\usepackage[T1]{fontenc}
\usepackage{graphicx}
\usepackage{subcaption}
\usepackage{float}
\usepackage{multirow}
\usepackage{ragged2e}
\usepackage{amsfonts}
\usepackage{nccmath}
\usepackage{tabularx, booktabs, makecell, caption}
\usepackage{siunitx}
\usepackage{caption}
\usepackage{subcaption}
\usepackage{hyperref}
\usepackage{tikz}
\usetikzlibrary{shapes,arrows,positioning}
\usepackage{algpseudocode}
\usepackage{algorithm}
\usepackage{comment}
\newcommand{\Cone}{C1 [L|L+R] }
\newcommand{\Ctwo}{C2 [R|L+R] }
\newcommand{\Cthree}{C3 [L+R|L+R] }
\newcommand{\Cfour}{C4 [L+R|R] }
\newcommand{\Cfive}{C5 [L+R|L] }

\tikzstyle{startstop} = [rectangle, rounded corners, minimum width=3cm, minimum height=1cm,text centered, draw=black, fill=red!30]
\tikzstyle{io} = [trapezium, trapezium left angle=70, trapezium right angle=110, minimum width=3cm, minimum height=1cm, text centered, draw=black, fill=blue!30]
\tikzstyle{process} = [rectangle, minimum width=3cm, minimum height=1cm, text centered,text width=4cm, draw=black, fill=orange!30]
\tikzstyle{decision} = [diamond, minimum width=3cm, minimum height=1cm, text centered, draw=black, fill=green!30]
\tikzstyle{arrow} = [thick,->,>=stealth]

\newenvironment{sciabstract}{%
\begin{quote} \bf}
{\end{quote}}

\newcounter{lastnote}

\title{Curriculum Is More Influential Than Haptic Information During Reinforcement Learning of Object Manipulation Against Gravity}
\author
{Pegah Ojaghi,$^{1+}$ Romina Mir,$^{2+}$ Ali Marjaninejad,$^{2,3}$Andrew Erwin,$^{2,4,5}$\\
	Michael Wehner,$^{6}$ Francisco J Valero-Cuevas$^{2,3,4,7,8\ast}$\\
\\
\normalsize{$^{1}$Computer Science and Engineering Department, University of California Santa Cruz,}\\
\normalsize{Santa Cruz, California, USA}\\
\normalsize{$^{2}$Department of Biomedical Engineering, University of Southern California, }\\
\normalsize{Los Angeles, California, USA}\\
\normalsize{$^{3}$Ming Hsieh Department of Electrical and Computer Engineering,}\\
\normalsize{University of Southern California, Los Angeles, California, USA}\\
\normalsize{$^{4}$Division of Biokinesiology and Physical Therapy, University of Southern California,}\\
\normalsize{Los Angeles, California, USA}\\
\normalsize{$^{5}$Mechanical and Materials Engineering Department, University of Cincinnati,}\\ \normalsize{Cincinnati, Ohio, USA}\\
\normalsize{$^{6}$Mechanical Engineering Department, University of Wisconsin-Madison,}\\
\normalsize{Madison, Wisconsin, USA}\\
\normalsize{$^{7}$Department of Aerospace \& Mechanical Engineering, University of Southern California, }\\
\normalsize{Los Angeles, CA, USA}\\
\normalsize{$^{8}$Department of Computer Science, University of Southern California,}\\
\normalsize{ Los Angeles, California, USA}\\
\normalsize{$^\ast$To whom correspondence should be addressed; E-mail:  valero@usc.edu.}\\
\normalsize{$+$These authors contributed equally to this work.}
}
\date{}

\begin{document} 


\baselineskip23pt

\maketitle

\newpage
% \begin{sciabstract}
% Learning to lift and rotate objects with the fingertips is necessary for autonomous in-hand dexterous manipulation. 
% A successful learning strategy depends on multiple factors such as the actuators and sensors available, the proposed control strategy, and the curriculum used. 
% While `curriculum learning' (sequentially appending task features) has been proposed as critical to learning of such multi-objective tasks, what such sequence is best for a given dexterous manipulation task remains debatable.
% It remains similarly debatable what haptic information at the fingertips best promotes such learning.
% We use model-free Reinforcement Learning to compare alternative curricula and haptic information (No-tactile vs. 3D-force sensing at the fingertips) when autonomously learning to lift and rotate a ball against gravity with a three-fingered simulated robotic hand with no visual input.
% We find that the choice of curriculum had the profound effect of biasing learning toward one or another combination of dexterous manipulation skills.
% Surprisingly, all curricula succeeded at learning even in the absence of tactile information.
% Lastly, we demonstrate our results generalize to balls of different weights and sizes.
% This work, therefore, challenges long-held notions about curriculum learning and the need for tactile information to autonomously learn in-hand dexterous manipulation.
% \end{sciabstract}
\begin{sciabstract}
Learning to lift and rotate objects with the fingertips is necessary for autonomous in-hand dexterous manipulation. 
In our study, we explore the impact of various factors on successful learning strategies for this task. Specifically, we investigate the role of curriculum learning and haptic feedback in enabling the learning of dexterous manipulation.
Using model-free Reinforcement Learning, we compare different curricula and two haptic information modalities (No-tactile vs. 3D-force sensing) for lifting and rotating a ball against gravity with a three-fingered simulated robotic hand with no visual input.
Note that our best results were obtained when we used a novel curriculum-based learning rate scheduler, which adjusts the linearly-decaying learning rate when the reward is changed as it accelerates convergence to higher rewards.
Our findings demonstrate that the choice of curriculum greatly biases the acquisition of different features of dexterous manipulation.
Surprisingly, successful learning can be achieved even in the absence of tactile feedback, challenging conventional assumptions about the necessity of haptic information for dexterous manipulation tasks. 
We demonstrate the generalizability of our results to balls of different weights and sizes, underscoring the robustness of our learning approach. 
This work, therefore, emphasizes the importance of the choice curriculum and challenges long-held notions about the need for tactile information to autonomously learn in-hand dexterous manipulation.

\end{sciabstract}

\section*{Introduction}

Dexterous manipulation is a triumph of biology \cite{lemon1995corticospinal, mackenzie1994grasping, valero2009hand, valero2017neuromechanical, billard2019trends, ortenzi2019robotic, liarokapis2016learning, sadun2016grasping}. 
However, the autonomous learning of such behavior continues to remain out of reach for robots \cite{andrychowicz2020learning, gupta2021reset, valero2017neuromechanical, kudithipudi2022biological, chen2023visual}. 
Robots have excelled at \textit{grasping} (reaching for and statically coupling an object to the hand by applying forces with the fingertips, fingers, and palm \cite{valero2017neuromechanical, ortenzi2019robotic, katyara2020leveraging, bicchi2000hands}) for decades (e.g., \cite{bicchi2000robotic, deimel2016novel, brown2010universal, murray1994mathematical, miller2004graspit, dollar2007sdm, mason1988planning, zeissler2022robotic, triantafyllidis2023hybrid}), but grasp is not dexterous manipulation \cite{valero2017neuromechanical}. 
\textit{Dexterous in-hand manipulation} (i.e., dynamically holding and reorienting an object with the fingertips \cite{valero2017neuromechanical, murray1994mathematical, cutkosky1990human, caggiano2023myochallenge}) is critical for interaction with, and use of, objects in unstructured human environments. 

To achieve this kind of manipulation with multi-fingered robotic hands, the robotics community has developed sophisticated control theoretical approaches \footnote{Henceforth we use the shorthand \textit{manipulation} to mean dexterous in-hand manipulation} (e.g., \cite{liarokapis2016learning, guo2017robotic, murray2017mathematical, valero2017neuromechanical, morgan2022complex, ambrose2000robonaut, catalano2014adaptive, bridges2010revolutionizing, castellini2009surface, huber2002robust}).
These control theoretical approaches, however, tend to require accurate models and state estimation, have narrow stability margins, and have difficulty compensating for friction, interpreting intermittent/deformable contact, and coordinating between multiple fingers.
As an alternative approach, biorobotic, neuromechanics, and artificial intelligence communities have introduced a variety of bio-inspired and data-driven machine-learning approaches (for reviews see~\cite{valero2009computational, valero2017neuromechanical, kudithipudi2022biological, bicchi2000hands, todorov2017learning}) in simulation and hardware.

One particularly promising approach is the sub-field of Reinforcement Learning (RL), which has provided several successful examples \cite{jeong2020learning,zhu2019dexterous,gupta2016learning, chen2023visual, marjaninejad2019autonomous, triantafyllidis2023hybrid}. 
RL empowers robots to iteratively enhance their manipulation skills through trial and error (without of a need of an accurate model of the task or the environment), resulting in gradual improvements within complex environments.
However, manipulation RL studies to date are usually highly computationally intensive---and have relied on vision---which limits their applicability  \cite{zhang2023piezo, todorov2017learning,kumar2015mujoco,funk2021benchmarking,cruciani2020benchmarking,morgan2022complex,theodorou2011neuromuscular,sun2022soft,thor2022versatile, lee2020learning,schilling2020modular,clune2011performance}. 
Lastly, most studies have been limited to the upward-facing hand configuration, relying on the palm as a resting platform for the object being manipulated which makes it an inherently more stable task to handle than a down-facing hand configuration\cite{andrychowicz2020learning}.
Adding the downward-facing hand configuration broadens the scope of solutions, delivering valuable insight to the robot manipulation community \cite{chen2023visual, akkaya2019solving, andrychowicz2020learning}.
However, it introduces additional challenges as this orientation requires the hand to counteract gravity at all times \cite{hu2023dexterous}, and errors can lead to instabilities and failure by dropping the object.
Here we use an RL based on the Proximal Policy Optimization (PPO)~\cite{schulman2017proximal} algorithm to autonomously learn manipulation with a downward-facing hand without direct vision.
%For two types of tactile information (No-tactile or 3D-force at the fingertips), 
We find that the choice of curriculum biases learning manipulation toward one or another combination of skills (i.e., lifting the ball and/or rotating it) more profoundly than the availability of tactile information.
 
Surprisingly, %curriculum was influential to the point that 
the absence of tactile information did not necessarily prevent or significantly degrade learning relative to the influence of curriculum.
These results reveal fundamental and previously underappreciated aspects of curricula as a powerful tool for autonomous learning of multi-objective tasks. 
For example, curricula commencing with both lift and rotation exhibit initial superior performance compared to those building up from simpler blocks, such as focusing solely on lift or rotation. 
Focusing on a single skill thereafter, however, can be additionally beneficial.
Beyond assessing the impact of curricula on autonomous manipulation, our study yielded the significant revelation that, contrary to long-held notions, the absence of tactile information (and direct vision) does not inherently impede or degrade the learning process. 
In fact, there seems to be a functional interaction with a curriculum where available sensing capabilities bias the learning process toward combinations of dexterous manipulation skills that can leverage the available tactile information.

\renewcommand{\thesubfigure}{\Alph{subfigure}}

\begin{figure}[ht!]
  \centering
     \begin{subfigure}{0.7\textwidth}
        \includegraphics[width=\textwidth]{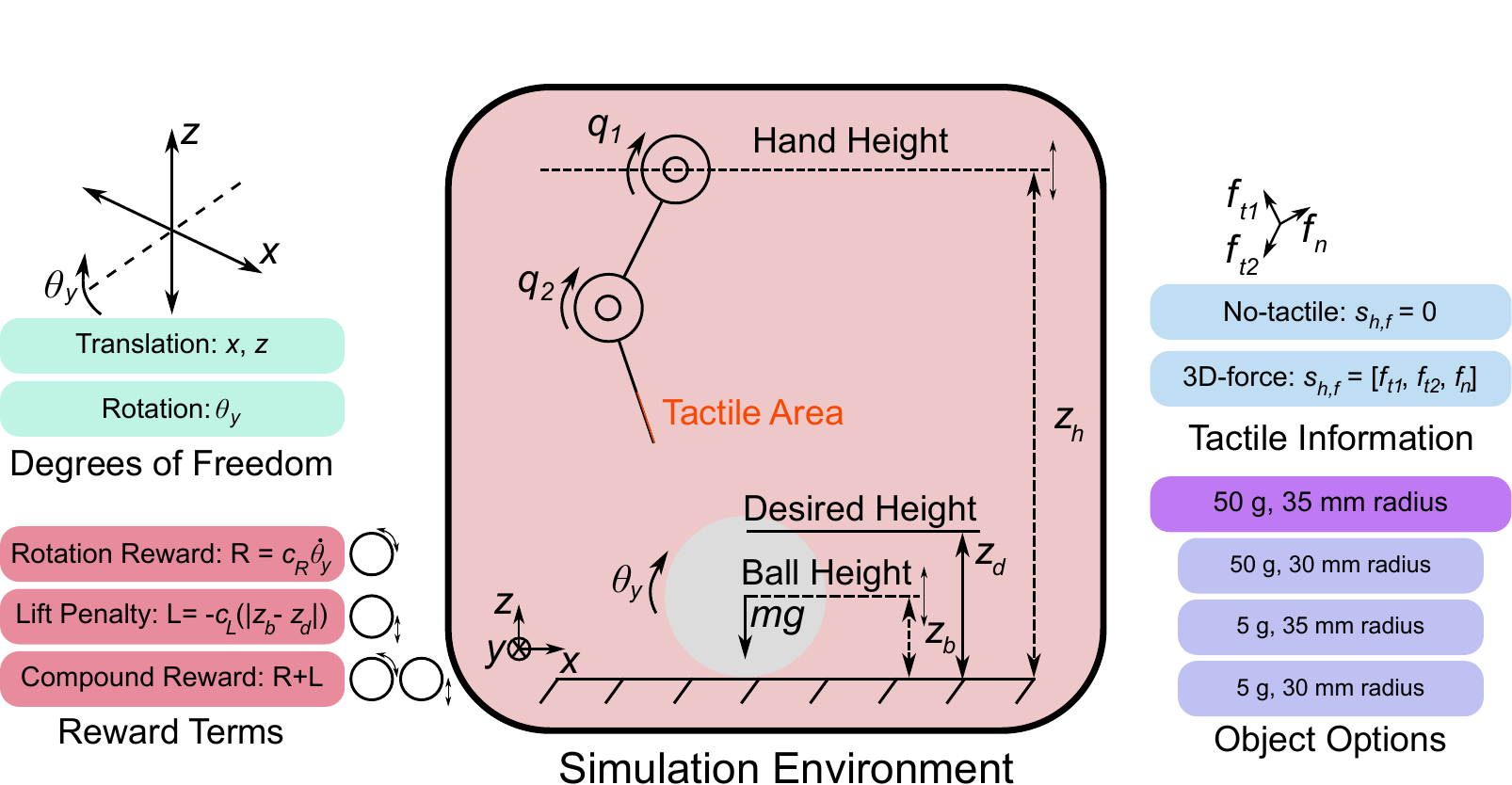}
        \caption{\justifying {\small }}
        \label{subfig:overview simulation}
    \end{subfigure}
    \quad
   \begin{subfigure}{0.88\textwidth}
     \includegraphics[width=\textwidth]{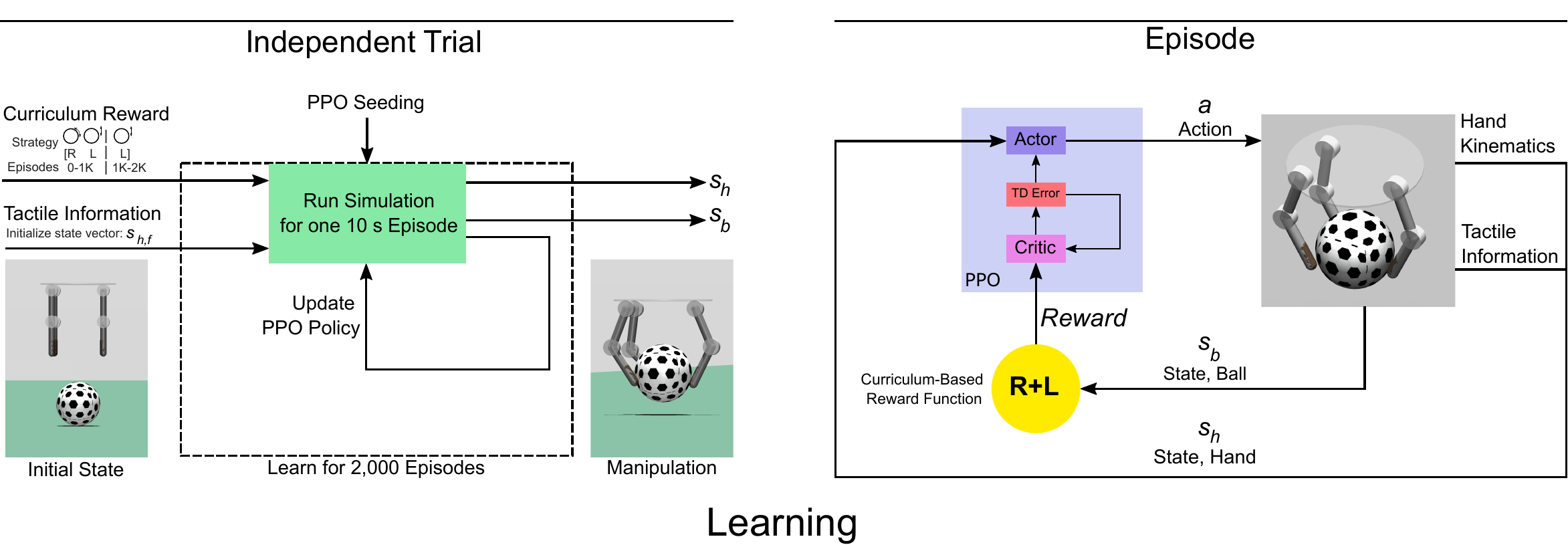}
        \caption{\justifying {\small }}
        \label{subfig_overview learning}
    \end{subfigure}
%general: 3 finger hand in MuJoCo
  \caption{\footnotesize{\textbf{Overview of Simulation Environment and Learning.}
  High-level overview of the simulation environment and learning approach to autonomous manipulation. See the Methods section for further details.
  \textbf{A: Simulation Environment}. A simulated three-finger robotic hand attempted to lift and rotate (i.e., dexterously manipulate) a ball. The 3D movement of the ball was lightly constrained to the X-Z plane. Changes in the ball state affect the reward, which is a function of rotation, lift, and/or a combination of the two. We tested this approach with two different tactile information conditions (No-tactile and 3D-force) available at the fingertips and four balls of different weights and sizes.  
  \textbf{B: Learning Algorithm.} 
  \textit{ Independent Trial, Left}: For each of the five curricula, autonomous learning was evaluated over 60 independent trials (one trial shown). Each trial in a curriculum consisted of two learning phases lasting 1,000 episodes for a total of 2,000 episodes. The reward function changed at the end of the first learning phase (with the exception of Curriculum 3, see Table \ref{tab:curricula}). 
\textit{Episode, Right:} Each episode lasted 10 s and began \textit{de novo} with the ball on the ground with the hand and fingertips suspended above it. In each episode, the PPO learning algorithm dynamically updates the agent's action 
(i.e., moving the fingers and hand) to increase the curriculum's reward.
}}
\label{fig: overview}   
\end{figure}

\section*{Results}
The goal of this project was to utilize curriculum-based RL to learn in-hand manipulation of an object against gravity in a data-efficient way---even while not using visual information.
We demonstrate how the choice of curriculum is more influential than tactile information when learning to lift and rotate a \textbf{ball (weighing 50 g with 35 mm radius)} with a three-finger robotic hand in simulation (Fig. \ref{fig: overview}). 
To do so, we systematically explored \textbf{two tactile conditions:} No-tactile (no force perception at all at the fingertip) vs. 3D-force (a 3D force vector in the direction of force at the fingertip) during  \textbf{five distinct curricula} (details in Methods). 
%For each of these 10 combinations (2 conditions x 5 curricula), we ran 60 independent replicates (i.e., \textbf{trials}) lasting 2,000 \textbf{episodes} (each lasting  10-s).
We defined each curriculum as implementing a learning policy that rewards various combinations and sequences of lift (\textbf{L}) and rotation (\textbf{R}) of a ball, which can switch at the halfway point (Methods, Table \ref{tab:curricula}). For example, Curriculum 1 (i.e., C1) only rewards lift (L) in the first half of the trial, and both lift and rotation (L+R) in the second half are described as [L|L+R]). 
% To determine the essentiality of tactile information, we conducted simulations under two distinct tactile sensory conditions: one incorporating 3D-force sensing at the fingertips, and the other involving the absence of tactile information at the fingertips (No-tactile).
% We find that 3D-force tactile information at the fingertips was useful---but not strictly necessary or always led to the best performance (Figs. \ref{fig: result_overview}, \ref{fig: overall_jointplot}).
Lastly, we confirmed the generalizability of our approach by learning to manipulate balls (objects) of different weights and sizes (Fig. \ref{fig: overal_violin}). 
We find that the order of reward (curriculum) greatly affects the progression of learning and the final performance, 3D tactile information was not consistently better than No-tactile information, and a similar trend was observed across all configurations (see the video file in Supplementary Information).

\subsection*{Curriculum profoundly affects the progression of learning and final performance}

\begin{figure*}[t!]
% \vskip 0.2in
% \begin{center}
\centerline{\includegraphics[width=1.0
\linewidth]{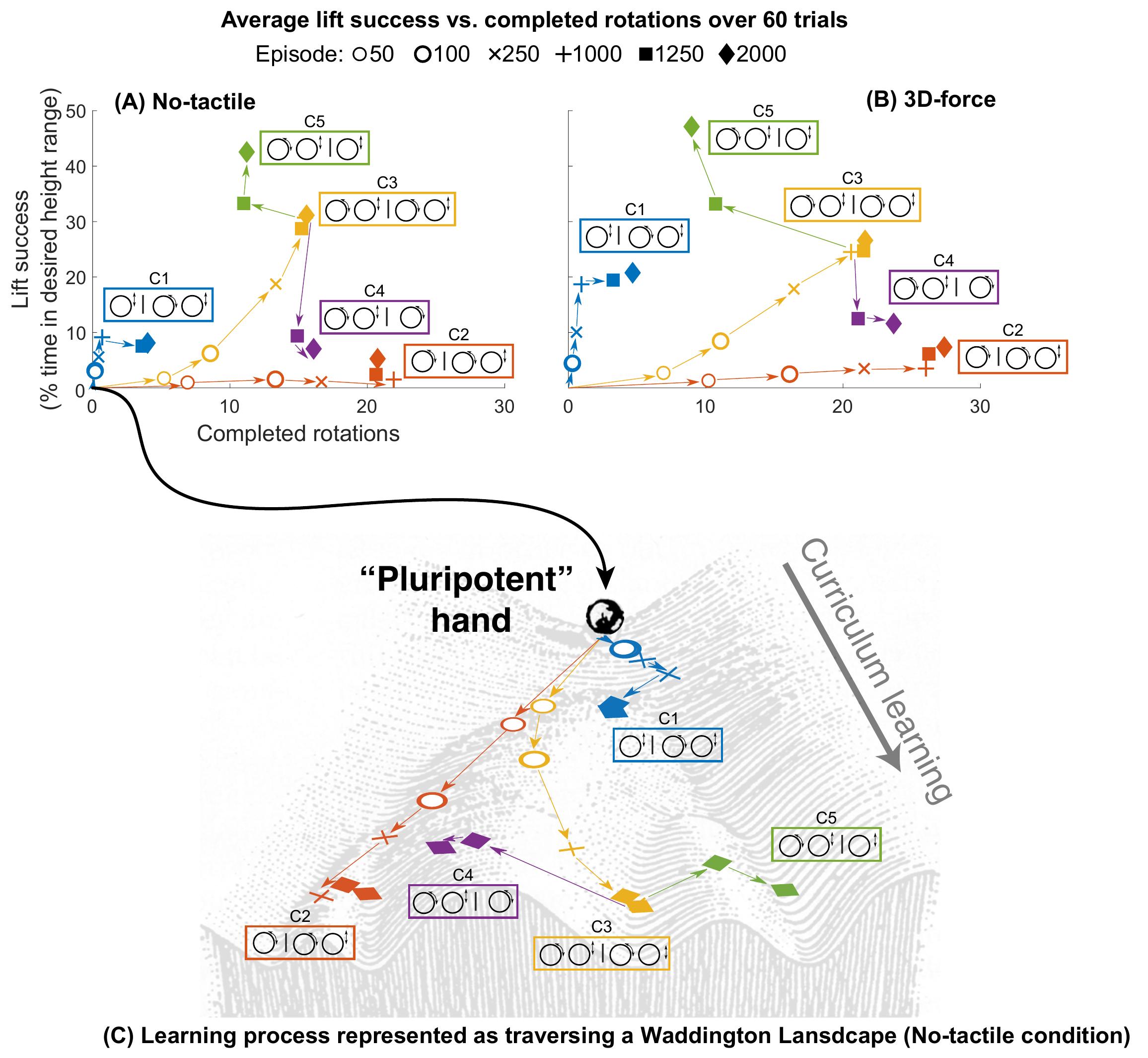}}
    \caption[]
    {\footnotesize \textbf{The evolution of learning highlights the dynamic functional interaction between curriculum and tactile information.}
    Manipulation performance during the last 10s of each episode noted: the percent of the time the ball is within the desired height range vs. number of complete rotations. Each point is the average of 60 independent trials. Arrows point in the direction of increasing episodes. Negative rotations were set to zero.
Note that the choice of curriculum had a profound effect on learning for both tactile conditions (\textbf{(A)} No-tactile and \textbf{(B)} 3D-force ).
 Surprisingly, learning happened even in the absence of tactile information, and manipulation performance was not always better with 3D-force information.
 \textbf{(C)} An analogy of learning as a developmental trajectory from a pluripotent state based on experience (curriculum). This effect of curriculum (and tactile information, cf. A vs. B) affects both learning (path) and final performance (endpoint), and can be visualized as traversing a `Waddington Landscape' (adapted from \cite{waddington1959evolutionary}).
   }
    \label{fig: result_overview}
\end{figure*}

Each combination of curriculum and tactile information (Methods, Table \ref{tab:curricula}) leads to a distinct evolution of learning and final performance. This effect of curriculum affects both the progression of learning (path) and final performance (endpoint), and can be visualized as traversing a developmental process (as ‘Waddington Landscapes’ in biology, Fig. \ref{fig: result_overview}, see Discussion).

Curricula, as expected, diverge in their ability to lift and rotate the ball. In fact, they had the profound effect of biasing toward one or another combination of skills (L or R) and also adapt to the available sensory input, much like experience-dependent developmental paths from an initial pluripotent state (Fig. \ref{fig: result_overview}C).
As we describe in detail in the Discussion section, we explicitly explored \textit{different} initial rewards with \textit{similar} final rewards (\Cone vs. \Ctwo), and vice versa (\Cfour vs. \Cfive).
In all cases,  the system was able to respond to the change in reward (albeit with variable success).
Note the evolution of skills for each curriculum tended to saturate quickly within the first 250 episodes of the first and second phases of learning. They tended to asymptote between the 250 and 1,000, and between the 1,250 and 2,000 episodes, respectively. 
Nevertheless, the final endpoints for each curriculum differed significantly, showing that curricula are more than simply a means to learn multi-objective tasks, but can actually produce different learning paths and endpoints---which can be exploited by the user to achieve different capabilities with the same na\"ive system (Fig. \ref{fig: result_overview}).

%1b
Counterintuitively, starting with a multi-objective reward can be as effective, if not more effective, than starting with simpler rewards.
% C3 better than C2  (both L and R) 
For example, rewarding \textit{both lift and rotation} during the first 1,000 episodes (\Cthree, \Cfour, and \Cfive) improves rotating the ball at the end of learning (episode 2,000) better than when only rewarding rotation (\Ctwo) at the start.

%Conrad Waddington's epigenetic landscape. The ball represents a cell and the bifurcating system of valleys represents trajectories of cell state. This diagram by C.H. Waddington neatly encapsulates the developmental pathways and progressive divergence of cells as they differentiate in the embryo. Reproduced from Waddington, CH © (1957) George Allen and Unwin (London).  

\subsection*{Tactile information is not necessary but can affect learning}
\begin{figure*}[ht!]
% \vskip 0.2in
% \begin{center}
\centerline{\includegraphics[width=1\linewidth]{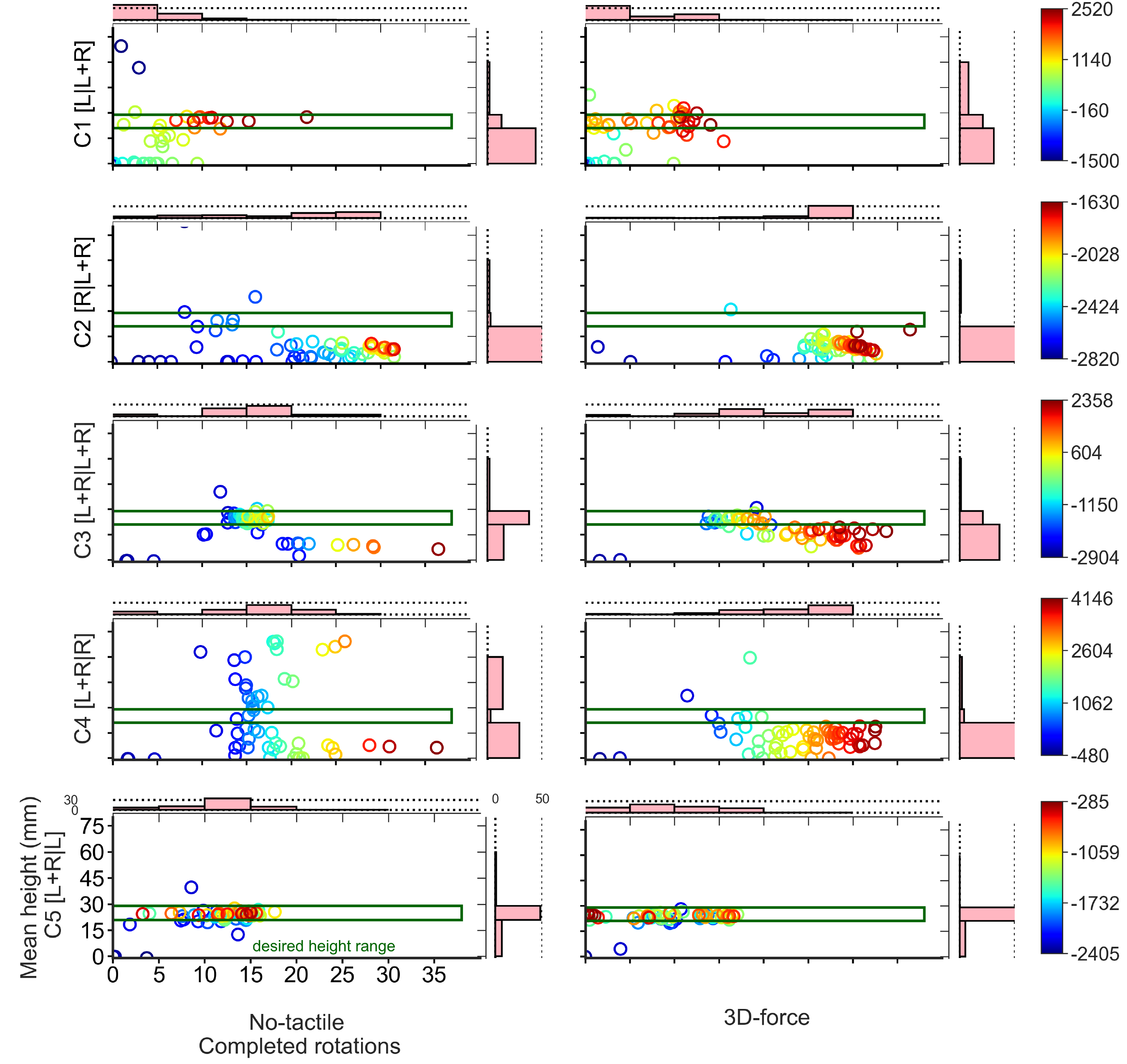}}
    \caption[]
    {\footnotesize \textbf{Performance across all curricula and both tactile information conditions.} The joint distribution illustrates the performance during the final 10s episode of each of the 60 trial runs (showcasing the mean ball height (mm) versus the number of completed rotations).
    The color-coded cumulative reward for the last episode of each run (refer to equation (\ref{eq:reward})) corresponds to different curricula. 
    Note that the final manipulation performance is represented by those points inside the green box defining the desired ball height (25 $\pm$ 4 mm). 
    } 
    \label{fig: overall_jointplot}
\end{figure*}

\begin{figure*}[t] %tp
% \vskip 0.2in
% \begin{center}
\centerline{\includegraphics[width=0.8\linewidth]{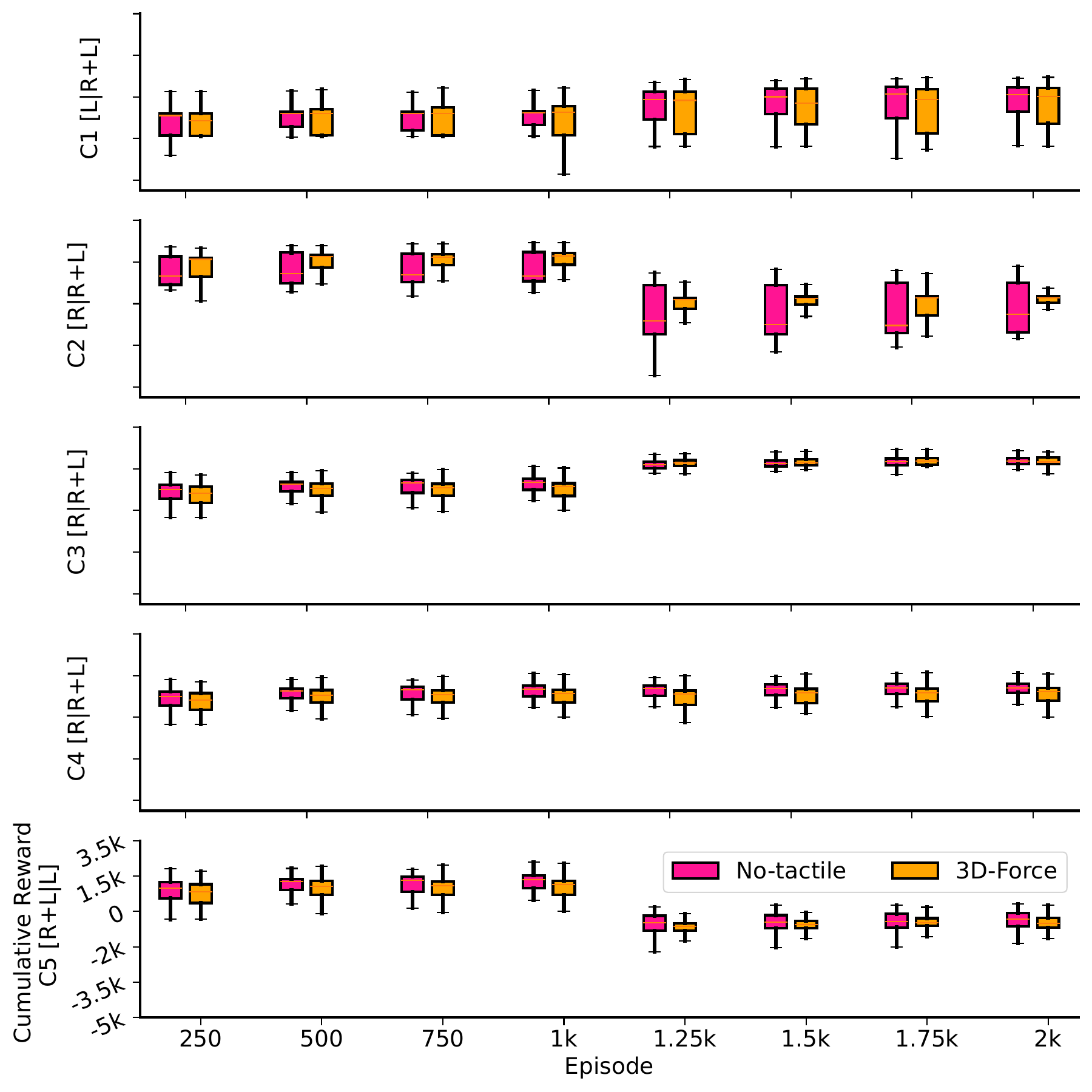}}
    \caption
    {\footnotesize \textbf{Cumulative reward across all curricula and tactile information conditions}. Boxplots, with median, across tactile conditions for 60 runs,  every 250 episodes. Note learning tends to saturate early.
    }     
    \label{fig: overall_boxplot}
\end{figure*}

\begin{figure*}[th!]
% \vskip 0.2in
% \begin{center}
\centerline{\includegraphics[width=1.02\linewidth]{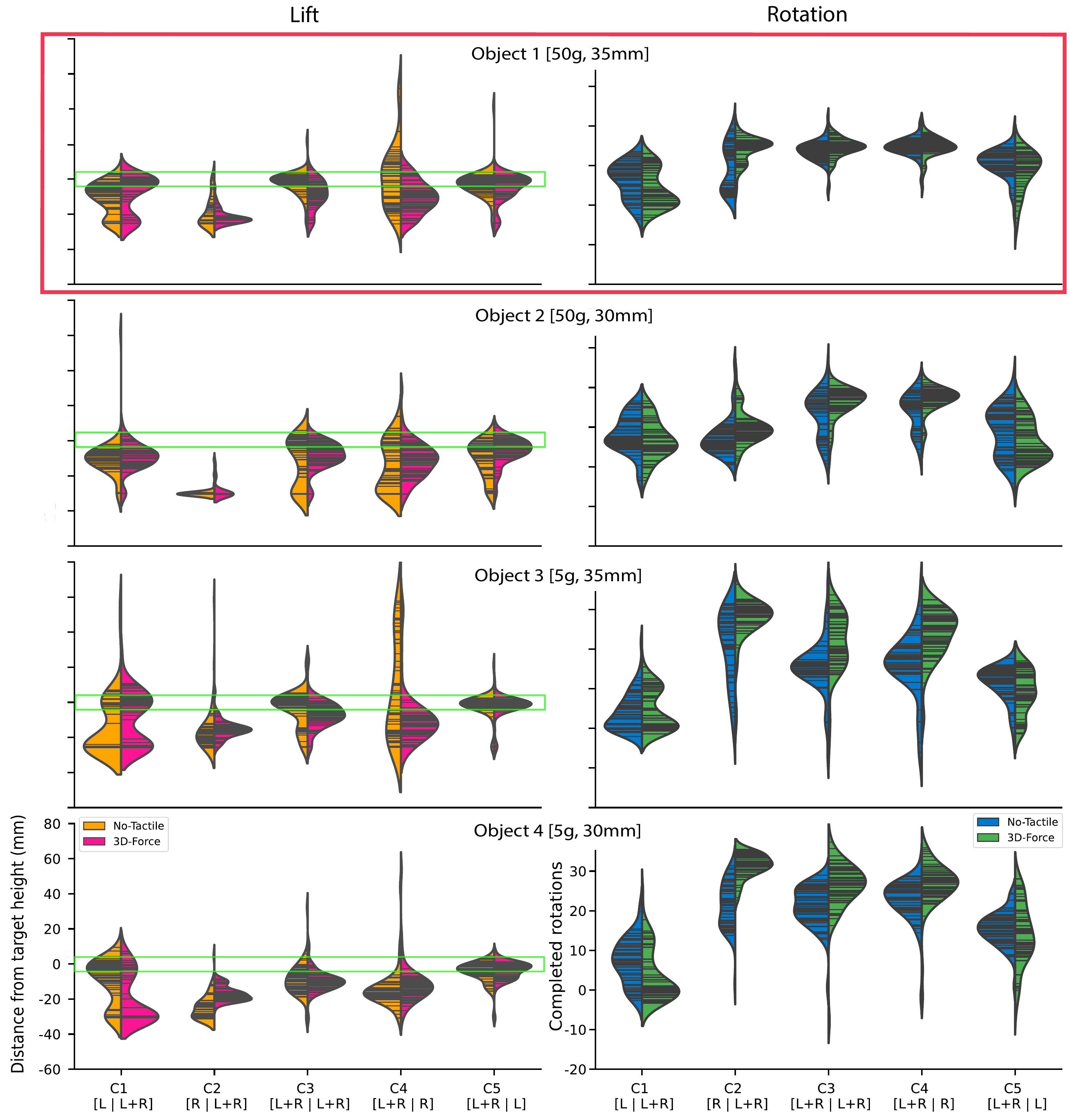}}
% {Figures/overall}}
    \caption[]
    {\footnotesize 
    \textbf{Final Performance for Lift (Left) and Rotation (Right) for both tactile conditions for four objects.} 
    The top row corresponds to the baseline object described in the main results. Violin plots show the distribution of Lift and Rotation at the end of learning (i.e., the last 10 seconds of the 2,000th episode) for all 60 trials. Lift is described as a distance from the desired height (the green box shows the distance from the desired height range $\pm$ 4 mm) and Rotation as the number of completed rotations for both tactile conditions, No-tactile and 3D-force.
    } 
    
    \label{fig: overal_violin}
\end{figure*}

% no-tactile learned
%3D has SOME effect, but often not much, and it makes the hand more complex. (some cases)
Most surprisingly, the absence of tactile information did not preclude learning. 
Moreover, learning with No-tactile information was comparable to the 3D-force information (Fig. \ref{fig: result_overview}).
The presence or absence of 3D-force information did, however, change the learning paths and endpoints of each curriculum (Fig. \ref{fig: result_overview}, \ref{fig: overall_jointplot})---but the effect was not uniform.
For example, 3D-force information did produced more lifting than No-tactile in \Cone at the end of learning. 
But this was reversed in \Cthree; and tactile information did not affect \Cfour or \Cfive much (Fig. \ref{fig: result_overview}).
This nuanced effect of tactile information at the end of learning is also seen in Fig. \ref{fig: overall_jointplot}, and on average during learning in Fig. \ref{fig: overall_boxplot}. 
This interaction was also seen while learning with different objects (see details in the Generalizability section and Fig. \ref{fig: overal_violin}).

% C5 is better lifter than C3  
Further nuance of the effect of tactile information can be seen in the different paths of learning and in the response to switching of rewards between the first and second learning phases (i.e., after episode 1,000). 
Note \Cthree rewards both skills during the entirety of both phases, but tends to be most effective at lifting in the No-tactile condition compared to 3D-force condition, Fig. \ref{fig: result_overview}.
Nevertheless, when switching the reward to only lift \Cfive or only rotation \Cfour at the end of the first learning phase, the 3D-force case makes up for lost ground and has endpoints similar to those for the No-tactile condition.
This effect seems to be reversed for \Cone and \Ctwo where only lift or rotation were rewarded at first. In these cases, the 3D-force condition produced greater lift and rotation during both learning phases.

%  The primary contribution of our work is how learning strategy influences manipulation performance.

% Discussion main points to be made:
% A. What did we learn about learning to manipulate?
%     Hard problem
%     Moving beyond curriculum learning to  a developmental approach to learning (waddington). 
%     Path vs. endpoint, we were able to continue to learn in interesting ways after the reward was changed (see C1 [L|L+R], C2 [R|L+R], C4 %1. Learning paths -- waddington plot --C1 and C2 comparison
    
% B. The role of sensory information
%     No need for tactile information or vision
%     tactile information alters curricula
%     no vision
    
% C. What did we learn about learning???
%     1. Beyond local minima: path vs. endpoint: the path is defined early on but system remains responsive did get `trapped' in a local minimum,  C2, C4
%      2. Capacity saturation -- C3
%     3. Transfer, no forgetting:  Interestingly, C4 [R+L|R] is not rewarded for lifting at the end of its curriculum but occasionally + C5lifts. 
% [No catstrophoc forgetting]
%     Multi-objective first, and then specialize?[C4, C5]
%     saturation in learning --
%     Generalizablity
    
% D. Comparison to the state of the art 
%     Need to implement scheduling of learning rate in PPO
%     learning rates
% E. Limitations, opportunities and future directions

\section*{Discussion}
\subsection*{What did we learn about learning to manipulate?}
\vspace{-30pt}
\subsubsection*\normalfont\textit{We provide proof-of-principle that it is possible to learn the hard problem of dynamic dexterous manipulation.} Putting our work in context is critical and best done by pointing to its place in the updated taxonomy of hand function put forth by MacKenzie, Iberall, Brand, Curkosky, Dollar, and others \cite{bullock2011classifying,feix2015grasp,mackenzie1994grasping,murray1994mathematical,cutkosky1989grasp, brand2012clinical}. 
In particular
% ---and as per the taxonomy of hand function put forth by Dollar and colleagues \cite{bullock2011classifying,feix2015grasp}---
we have addressed the problem of dynamic manipulation with three fingers while the ball is at risk of being dropped at any moment (see `Comparison to State-of-the-Art' section).
This definition emphasizes that `grasp' and `pick-and-place manipulation' are conceptually and mechanically distinct from `dynamic manipulation' as addressed here, even though they are at times used interchangeably in the literature \cite{valero2009whyhand}.
Such dynamic manipulation is, in fact, an enviable ability that is also difficult for biology to achieve as it develops in humans late in childhood, degrades in healthy aging, and is quickly lost in even mild/initial forms of neurological conditions such as peripheral neuropathies, stroke, and Parkinson's disease (e.g., \cite{duruoz2020hand,valero2009whyhand}). 
In our work, the fingertips induce dynamic translation and rotation of the ball while making and breaking contact. 
As such, the hand function we achieved merits the description of dynamic dexterous manipulation.
% \vspace{-30pt}
\vspace{-2em}
\subsubsection*\normalfont\textit{Curriculum learning can be seen as a developmental process from a pluripotent state.}
We use the analogy of the Waddington Landscape (Fig. \ref{fig: result_overview}C) for curriculum learning of manipulation because of its similarity to epigenetic transformation from a pluripotent state in biological development \cite{waddington1959evolutionary,guerrero2022epigenetics}. Curriculum learning, in fact, produces a developmental trajectory from a nai\"ve (i.e., pluripotent) state based on resources (tactile information) and experience (curriculum) (Fig. \ref{fig: result_overview}A\&B).
Each curriculum affects both the progression of the learning (path) and its final performance (endpoint), and can thus be thought of as traversing a Waddington Landscape.
%This visualization of cell type differentiation \cite{arthur2010evolution, hall2004search} is analogous to how curriculum learning produces a specific, experience-driven path that `channels' abilities to a particular final skill set.
Importantly, the evolution of skills for each curriculum was (unlike cell differentiation) not strictly irreversible, but remained adaptable.
% Curricula did not get `trapped' in a local minimum, but were able to continue to learn in interesting ways after the reward was changed (see C1 [L|L+R], C2 [R|L+R], C4 [L+R|R] and C5 [L+R|L] in Fig. \ref{fig: result_overview}).
% Specifically, note that the change of reward after the first learning phase caused an initial learning mostly exclusive to the new skill (see ~90\textdegree shifts in paths in Fig. \ref{fig: result_overview}).
Specifically, the change of reward after the first learning phase did not preclude the system from emphasizing the improvement of the new skill. This is visually represented by ~90\textdegree{} shifts in the paths (see  \Cone and \Ctwo in Fig. \ref{fig: result_overview}).
In some cases, the response to a switch in reward even reversed a learned skill for the first 250 episodes in the second phase of learning, and only then increased the new skill (see \Cfour and \Cfive in Fig. \ref{fig: result_overview}). 
In one case, \Cthree, there was no change in reward after the end of the first learning phase, and the system was saturated already.
In others, the system did respond like an `irreversible' system that learned little of the new skill, of at all,  when the reward function was switched (e.g., \Ctwo in the 3D-tactile case in Fig. \ref{fig: result_overview}). See the next Discussion Section.

\subsection*{The role of sensory information} 
\vspace{-30pt}

\subsubsection*\normalfont\textit{Manipulation can be achieved without tactile information or vision.}
Tactile information has long been thought as necessary for human---and by extension robotic---manipulation \cite{valero2017neuromechanical,wettels2008biomimetic}.
This idea was reinforced by the work of Johansson and Westling \cite{johansson1984roles,johansson1992somatosensory} demonstrating that numbing the fingerpads with temporary anesthetic greatly impairs fine manipulation.
Our results in Fig. \ref{fig: result_overview} provide a counter-example to this longstanding notion.
Interestingly, we found that our system was able to learn even in the absence of tactile information (the No-tactile condition in Figs. \ref{fig: result_overview}A and \ref{fig: overall_jointplot}).
In fact, having 3D-tactile information not always produced better performance (cf. \Cthree in Fig. \ref{fig: result_overview}A\&B).

% The answer, we believe, comes from the nature of the manipulation task. As described in Fig. \ref{fig:overview}B,  PPO, as a reinforcement algorithm, is simply conditioning actions (i.e., next-step finger joint angles and hand height) on the state of the system to maximize the reward. 
% The state of the system consisted of finger joint angles and angular velocities in the No-tactile case, and finger joint angles and angular velocities \textit{plus} 3D-force at the fingertips in the 3D-tactile case.For details see Methods. 
How is it possible to learn to manipulate without vision or tactile information? 
The answer, we believe, comes from the nature of reinforcement itself. 
As described in Fig. \ref{fig: overview}B, PPO---as a reinforcement learning algorithm---condition its actions (next-step finger joint angles, angular velocities, palm position, and velocity) based on the system state, ultimately optimizing for increased reward.
In the No-tactile case, the hand's state comprises finger joint angles, angular velocities, palm position, and palm velocity---which seem to suffice to learn the task. 
% In the 3D-tactile condition, the representation of the hand's state expanded to encompass not only finger joint angles and angular velocities but also includes 3D-force data at the fingertips (for details see `Methods' section)---which also sufficed to learn the task.
Therefore, lift and rotation of the ball was a product of guided hand kinematics that properly affect ball dynamics to increase the reward in the No-tactile case, such as in our previous work to learn locomotor movements without the need to sense the ground \cite{marjaninejad2019autonomous}.
% and that of others giving rise to the field of data-driven robotics \cite{kawaharazuka2022learning, yang2020data, allam2023data}.
% Importantly, the number of fingers (three) and their possible configurations sufficed to induce ball dynamics that drove the reward. 
%In these cases, there is no explicit need for a control law (or at least a formal pre-defined model) to produce the desired behavior.

%
% While some may question whether our work is `manipulation,' we believe it is as much as juggling or walking is in those other examples.
% Moreover, this ability to produce useful behavior---including manipulation---by mapping states to actions is one of the benefits and core capabilities of reinforcement learning \cite{schulman2017proximal,wang2020reinforcement, andrychowicz2016learning,andrychowicz2020learning,hu2023dexterous}. 
As such, a main contribution of our work is to provide an existence proof that an agent using reinforcement learning is able to learn a sophisticated manipulation behavior even in the absence of tactile information.
Note that direct vision was not necessary either, as in other prior work \cite{marjaninejad2019autonomous}.
%(i.e., blind juggler, passive walker and our manipulator).
%This is akin to the `blind juggler' D'Andrea created \cite{reist2012design} or passive dynamic walkers by McGeer \cite{mcgeer1990passive}.
Our important result about dynamic manipulation provides impetus to revise our thinking about, and use of, tactile and visual information to allow freer thinking for engineers (and bio-roboticists) creating the next generation of dexterous hands and robots. 

\vspace{-2em}
\subsubsection*\normalfont\textit{The presence or absence of tactile information did, however, alter the progression of learning.}
Figures \ref{fig: result_overview}A\&B (and the Supplementary Information Fig. \ref{fig: all_overview}) show that the type of sensory information did affect learning---but the general features of the path and endpoint of each curriculum remained similar in both tactile conditions.
Importantly, the effect of tactile information was not systematic. The 3D-tactile cases were not consistently or necessarily better than the No-tactile cases, or vice versa.
Thus, curriculum is a dominant factor compared to tactile information.

%One possibility is that the PPO algorithm performs slightly differently when the length of the hand state vector $s_h$ is extended by 12 elements (3 forces per finger).
From the computational perspective, one could have expected that when learning with a fixed number of episodes, 3D-tactile information would perform systematically worse because of the computational demands associated with extending the length of the hand state vector $s_h$ by 9 elements (3 forces per finger) for the same PPO algorithm architecture which now has to tune more weights (Fig. \ref{fig: overview}).
But, 3D-tactile cases at times outperformed the No-tactile cases (e.g., \Cone in Fig. \ref{fig: overview}A\&B), which strongly suggests that our comparisons across curricula and tactile conditions are not the result of an imbalance in computational demands for a fixed number of learning episodes (1,000 per learning phase for a total of 2,000).
This is additionally supported by the fact that 3D-tactile cases also saturated their learning by the $250^{th}$ episode (like the No-tactile cases).

Also, it is important to note that this study does not undermine the effectiveness of tactile information in many everyday tasks. 
It merely provides a proof-of-principle that it is possible to learn a specific task (i.e., the manipulation task of interest in this paper)  without using tactile information; and with performance comparable to when tactile information is available. 
It is clear that many tasks exist for which sensory signals would either be crucial to perform, or would greatly enhance, either the learning speed for the task, the final performance, error correction, and/or their robustness and repeatability. These are beyond the scope of our work.

\subsection*{What did we learn about learning?}

\vspace{-30pt}
\subsubsection*\normalfont\textit{Our system exhibits some important features of lifelong learning.}
As defined in \cite{kudithipudi2022biological}, our system shows \textit{transfer and adaptation} because it reuses knowledge to improve performance and rapidly adapts to novel skills as in \Cone\Ctwo, \Cfour, and \Cfive (in Figs. \ref{fig: result_overview}\ref{fig: overall_jointplot}, and \ref{fig: overall_boxplot}).
% , overcomes catastrophic forgetting (i.e., retains previously learned knowledge while training on new tasks), 
Similarly,  our system did not suffer from \textit{catastrophic forgetting} as it was able to retain varying amounts of previously learned knowledge on a case-by-case basis (Fig. \ref{fig: result_overview} and Supplementary Information, Fig. \ref{fig: all_overview}).
For example \Cfour and \Cfive did not entirely forget to lift or rotate when they were no longer rewarded, respectively. 

\vspace{-2em}
\subsubsection*\normalfont\textit{Curriculum learning does not necessarily have to advance gradually from single-objective to multi-objective rewards.}
In many applications such as locomotion, investigators have found that curriculum learning is indispensable to advance gradually from single-objective (i.e., `simpler') to multi-objective (i.e., `more complex') rewards \cite{ramdya2023neuromechanics}.
This has led to curriculum learning becoming the standard approach in the field.
From the traditional definitions of Vanilla or Progressive curriculum learning \cite{bengio2009curriculum,soviany2022curriculum}, one might assume that first learning to lift the ball (a form of grasp) is 'easier' than rotating it, which involves a dynamic behavior \cite{valero2017neuromechanical,murray2017mathematical} and a curriculum strategy in which rotation is learned only after lift is going to be a significantly more successful one.
%However, the effectiveness of learning depends on how and when we introduce this new task.
%When both rotation and lift are rewarded exclusively during the initial half of the learning phase, a prolonged learning phases is not required to attain proficiency in manipulation.
However, rewarding lift \textit{and} rotation from the start does not hinder learning, as demonstrated by \Cthree.
In fact, it allowed transfer and adaptation for \Cfour and \Cfive to subsequently refine the single skill rewarded during the second phase of learning---albeit at the expense of some reduction of the non-rewarded skill. %(locations in  fig. \ref{fig: overall_jointplot}).
However, it is noteworthy that curricula that rewarded only one skill from the start (\Cone and \Ctwo) were not able to learn the second skill as efficiently during the second learning phase (rotation and lift, respectively).

%%% Capacity Saturation ---C3 
% Capacity saturation (
Another aspect of lifelong learning involves the saturation of capacity causing learning to slow down \cite{mccloskey1989catastrophic, sodhani2020toward}.
Capacity saturation arises due to the fixed representational capacity of parametric models, including the PPO algorithm \cite{sodhani2020toward}. 
We see this in our implementation of PPO---which increasingly fails to absorb additional knowledge from successive episodes. 
This is most evident in \Cthree for the entire second phase of learning, as shown in Fig. \ref{fig: result_overview}. 
A learning model with more free parameters would theoretically be able to absorb additional knowledge from successive episodes.
%But it is important to note that relatively more data would also be needed for such scenarios.

\vspace{-2em}
\subsubsection*\normalfont\textit{The curriculum-based learning rate scheduler enhances the efficiency of learning which accelerates convergence to higher reward}

We sought to align the implementation of learning rates in PPO with the nature of curriculum learning.
To do this, we defined our curriculum-based learning rate scheduler to adjust the linearly-decaying learning rate when the reward changed (Fig. \ref{fig:learningrate}).
We find this improved learning and allowed a more fair comparison across curricula as it reduced heuristic tuning efforts. 
% Therefore, our approach with the curriculum-based learning rate scheduler plays a crucial role in improving the acquisition of manipulation skills by efficiently adapting the learning process (Fig. \ref{fig:learningrate}).  
This curriculum-based learning rate scheduler offers an effective approach tailored to curriculum learning for autonomous systems by modifying the learning rate only when changing task complexities and rewards. 
%Our study highlights the significant impact of such adaptability, especially in robotic manipulation where rapid skill acquisition is crucial. 
Empowering curriculum learning to adapt learning rates in a way compatible with changing rewards enables autonomous systems to learn complex and dynamic environments more systematically, autonomously, and effectively. 
Thus, integrating curriculum-based learning and reward scheduling into a `curriculum-based learning rate scheduler' for autonomous systems is vital to enhance their learning capabilities and performance in manipulation tasks.

\vspace{-2em}
%%% Generazablity
\subsubsection*\normalfont\textit{Lastly, we demonstrate our results generalize to balls of different weights
and size.}
As shown in the Supplementary Information in Figs. \ref{fig: all_overview}--\ref{fig:S4_boxplots}, our results were consistent across the four objects we studied (i.e., of two masses, 50 and 5 g, and two sizes, 35 and 30  mm in radius, Fig. \ref{subfig:overview simulation}).
Namely, our system was successful at learning to manipulate, but in a way that curriculum had a greater impact than tactile information. 
There were minor differences across the endpoint performance for each object (note the difference is the scales of the axis).
But the learning paths for each curriculum, and the effect of switching the reward, remained consistent (Fig. \ref{fig: all_overview}). 
This can also be seen in the detailed depiction of the distribution of rewards as learning progressed, Fig. \ref{fig: overal_violin}).
This further shows that the effect of ball size or weight (like that of tactile information as mentioned above) was not substantial nor systematic.

%old:  This brings us to the essence of curriculum learning, a strategic method documented in the literature \cite{wang2021survey, elman1993learning, marjaninejad2021autonomous}. It involves the gradual training of a machine learning model, commencing with simpler subtasks and progressively advancing toward mastering the entire target task. This approach holds particular significance in reinforcement learning (RL), where the decision-making process of selecting an appropriate curriculum and defining the sequence of reward functions for acquiring increasingly complex skills plays a pivotal role \cite{bengio2009curriculum, pentina2015curriculum}."

% This exploration gains particular relevance given the advanced tactile sensing capabilities observed in humans, emphasizing the significance of touch in the design and control of robotic hands \cite{dahiya2009tactile, wettels2009multi, matulevich2013utility}. Crucially, we establish the generalizability of our approach to objects of diverse sizes and weights.

\subsection*{Comparison to the state of the art} 
It is critical to note that, as we have stated in the past \cite{valero2017neuromechanical}, \textit{grasp} or \textit{pick-and-place tasks} are not \textit{dexterous manipulation} in the rigorous sense of grasp taxonomies.
Even reorienting a cube resting on an upward-facing palm \cite{andrychowicz2020learning, deimel2016novel,yin2023rotating, qi2023general, shenfeld2023tgrl, yang2023tacgnn, akkaya2019solving} is not prehensile manipulation.
Likewise, Prior work has used extensive vision with the upright palm to hold an object being reoriented \cite{andrychowicz2016learning,chen2023visual}, other than one demonstration of learning under increasing force of gravity \cite{sievers2022learning}, we know of no published demonstration of dynamic manipulation task against full gravity utilizing curriculum learning either directly in simulation or hardware. 
We employ a novel curriculum-based learning rate scheduler for PPO, which significantly enhances the success performance across all scenarios.
We now discuss how our novel approach to manipulation compares and contrasts with other studies in robotics and RL.
The state-of-the-art of \textit{autonomous learning} for in-hand manipulation is limited. 
Although important advances have been made using computationally intensive approaches in simulation and hardware (e.g., \cite{todorov2017learning,kumar2015mujoco,funk2021benchmarking,cruciani2020benchmarking,morgan2022complex, chen2023visual}), these tend to be impractical for autonomous learning at the edge. 
Augmenting RL for manipulation with imitation learning has shown some successes \cite{jeong2020learning,zhu2019dexterous,gupta2016learning, chen2023visual}, but collecting task-specific expert demonstrations from humans are often limited to specific objects or tasks, might not always be practical, require specialized equipment and can be time-consuming.

In contrast, we used a model-free data-driven approach because precise prior knowledge of the system, objects and the environment is not always available, especially in unstructured environments. 
Although some other studies also use model-free RL methods for rotating objects with simulated fingers or a robotic hand \cite{van2015learning,melnik2019tactile,andrychowicz2020learning,chen2021system}, we have overcome some of their drawbacks.
% For example, \cite{van2015learning} explored rolling an object with the palm facing down in hardware, but the goal was not to lift the object and they used a tabletop to support the rolling. 
In \cite{melnik2019tactile,andrychowicz2020learning}, the orientation of an object was controlled while resting on an upward-facing palm. 
Thus, it did not have to be held against gravity as it was not at risk of being dropped at any time.

Some of these limitations were addressed by Chen \textit{et al.} \cite{chen2021system} in simulation by manipulating the object with the palm facing downward like we did, but gravity was introduced slowly as part of the curriculum.
Moreover, to successfully manipulate the object the authors found it important to `initialize the object in a stable configuration'---which we did not need.

The way our work went beyond the state-of-the-art, therefore, is by demonstrating for the first time a method with the ability to autonomously learn to manipulate an object against gravity while revealing the role of curriculum learning and tactile information in in-hand manipulation. 
% Moreover, it does so based on a data-driven approach that uses few shots (limited experience) with novel approach to learning rate. 
The impact of learning rate scheduling on stochastic optimizer performance has been extensively investigated in recent research \cite{cooper2021hyperparameter, xu2019learning}. In our study, we specifically explore the effects of a constant and linear piecewise learning rate for PPO on the success of our architecture.
After careful consideration, we have decided to proceed with the piecewise learning rate. This adaptive approach adjusts rates dynamically throughout training, speeding up the process with higher initial rates and ensuring stable convergence with lower rates later on.

% Importantly, our work focuses on manipulating the objects against gravity with the risk of being dropped at any time.
Lastly, our work underlines the importance of curricula in manipulation and shows how the right choice of a curriculum can enhance performance and robustness across multiple tasks by exhibiting some important features of lifelong learning.

\subsection*{Limitations, opportunities and future directions} 
While our work pushes the field of autonomous manipulation forward, it naturally has some limitations. 
First, our work is done in simulation.
But, as with many other studies looking to bridge the sim2real divide \cite{marjaninejad2019autonomous,ruppert2022learning}, we used realistic physical constraints within our state-of-art physics engine (MuJoCo) that handles dynamic contacts and impacts well. 
This is a foundation that will enable future implementations in hardware. 
As to the geometry of our hand, it is common for useful robotic hands to have three fingers \cite{van2015learning,morgan2022complex}. 
Curriculum learning has multiple varieties \cite{soviany2022curriculum} that can adapt as learning progresses such as Self-Paced curriculum learning.
In our case, our learning phases were of fixed duration even though the system tended to plateau.
Thus, it could benefit from future implementations that adapt reward changes to minimize training time. 
Lastly, our manipulation tasks serve as a foundation for---but do not yet address---traditional use cases for activities of everyday life. 

%% added by Parmita and reviewed by Ali on Feb 4th
Our choices regarding PPO, curriculum design, hand and object structure, reward function, and other parameters were specifically tailored to address the scientific questions of interest within the scope of this paper and to establish a proof of concept. It's important to emphasize that our selections were not intended as universally applicable solutions.
That is to say, to address a different need, a similar pipeline to this paper can be utilized but different tasks, environments, or robotic structures might need to be used. Also, different learning blocks (different than the RL technique or the adaptive curriculum-based learning rate scheduler function used in this paper) can be used that might serve best for another specific task or purpose.

\section*{Methods} \label{methods}

In this section, we first describe the simulation environment and the task used in this study. Then, we elaborate on the learning policy that enabled autonomous manipulation.

\subsection*{Simulation environment}
The manipulation and machine learning communities have used the advanced physics simulation environment MuJoCo \cite{todorov2012mujoco} for tasks involving autonomous manipulation.
MuJoCo allowed us to implement reinforcement learning algorithms on a robotic hand in a realistic environment that includes contact dynamics (including penetration) and gravitational acceleration \cite{todorov2012mujoco, kumar2014real}.

To demonstrate the adaptability and robustness of our proposed methodologies, we assessed the performance using four different objects.
Our evaluations encompassed systematic exploration, considering two different weight combinations (50 g and 5 g), as well as varying ball radii (35 mm and 30 mm). 
The work presented herein focuses on a ball of 50 g with a radius of 35 mm with the other configurations presented in the section Generalizability in Supplementary material.

\subsubsection*{\textit{Robotic Hand Design.}}

We simulated a bio-inspired, three-fingered robotic hand with a palm and three identical servo-driven fingers: two adjacent fingers, analogous to the 'index' and 'middle' fingers, and one opposing them, analogous to the 'thumb'. In contrast to our prior efforts \cite{mir2021active}, where we showcased the reach-to-manipulate capability with a downward-facing orientation using distinct curricula, we modified the hand design.
Each finger consisted of two joints that could rotate about the $y$-axis ($q_1$ and $q_2$ in (Fig. \ref{subfig:overview simulation})), similar to the flexion or extension seen in human fingers.
The size of the palm and length of each `phalanx' was based on an average human hand \cite{kumar2016optimal,van2015learning}. 
An additional servo motor was included at the base of the hand, which provides translational motion in the vertical direction ($z_h$). 

\subsubsection*{\textit{Fingertip Tactile Sensors.}}

This work incorporated tactile information and RL, sometimes referred to as touch-augmented RL, as we covered the internal side (i.e., the `pads' of the fingertips) of the distal phalanx of each finger with tactile sensors. 
Contact regions were configured near the tips of each finger (Tactile Area, Fig. \ref{fig: overview} ). Objects contacting the finger outside of these tactile areas (sites, in MuJoCo) are not perceived as tactile information by the learning algorithm \cite{todorov2012mujoco,melnik2019tactile}.

We used MuJoCo's built-in features to record the 3D-force sensor on the fingertips of all three fingers. 
The 3D-force sensor sites provide a 3D array of 3 orthogonal forces (one normal and two tangential to the sensor site for each sensor) of scalar values representing the 3D-force vector. 
Moreover, we have considered an additional case: No-tactile. In the No-tactile case, the state vector for the tactile information $s_{h,f}$ is null (we do not consider the tactile information in learning).
As shown in Fig. \ref{subfig:overview simulation}, the possible contact tactile information at each fingertip is indicated by~$\mathbf{s_h,f}=[f_{t,1}, f_{t,2}, f_{n}]$ and it depends on tactile sensing available at fingertips. See Supplementary Table \ref{tab:tactile force info} for more details on the tactile information.

\subsubsection*{\textit{Task Description}}

The robotic hand attempts to manipulate a 50 g, 35 mm radius ball, which starts each episode on the ground with the palm of the robotic hand at a height of 200 mm above the ground.
The ball height~$z_b$ is defined at the center of the ball, and we specified a desired height for the ball~$z_d$ to be ~$25$ mm above ~$z_b$. In other words, the desired height~$z_d$ is 60 mm above the ground. 
Through simulation constraints, the ball is limited to 2 translational DOFs (moving vertically~$z$ and horizontally~$x$) and 1 rotational DOF (rotation about the ~$\theta_y$ direction; see (Fig. \ref{fig: overview})). We included viscous damping in the translational and rotational DOFs of the ball to stabilize the simulation and prevent numerical instabilities for the simulation of the rigid fingers. 

We further limited the ball's movement in the~$x$ direction by adding stiffness to the ball. The details of the simulation parameters, including the robotic hand and the ball) are shown in Supplementary Table \ref{tab: hand_parameter}.

\subsubsection*{\textit{Observation and Action Space}}

The system's state vector includes the hand state vector ($\mathbf{s_h}$), consisting of fourteen kinematic degrees of freedom (DOFs), along with the position and velocities of the hand's palm ($\mathbf{s_p}$) (2 DOFs), and the position and velocity of the ball ($\mathbf{s_b}$) (6 DOFs). This 20-dimensional vector encapsulates joint angles ($q_1$ to $q_6$) and their derivatives, as well as the vertical height of the hand ($z_h$) and its derivative, collectively describing the dynamic state of the system.

Additionally, the ball state vector comprises vertical ($z$) and horizontal ($x$) translation, and its rotation about the $y$-axis ($\theta_y$). No other translations or rotations are permitted (see Supplementary Table \ref{tab: simulation parameters generalizability}). The height of the hand, $z_h$, is actuated for the hand to reach for and manipulate the state of the ball ($\mathbf{s_b}$) by rotating ($\theta_y$) and lifting ($z_b$) it to a desired height ($z_d$).

It's important to note that not all state variables are utilized in our reinforcement learning policy (observation state). Specifically, the observation state omits details about the ball's velocity and position, as explained in the following subsection. Furthermore, it's worth mentioning that the action space aligns with the observation state.
When a 3D-force is introduced, the state of the system dynamically changes, augmenting the hand state with an additional 9 data points.
% In any physical system or simulation environment, monitoring the position and velocity of a moving object (not in contact with the hand) would be performed via vision. 
% By excluding ball states  from our learning policy, we enable a learning system which does not require vision (because it does not assume perception of these external states).

\begin{algorithm}
\caption{Simulation with PPO}
\begin{algorithmic}[1]
\Procedure{RunSimulation}{}
\State \textbf{Initialize} simulation environment
\State Set random seed for reproducibility
\State Initialize policy and value networks
\State Initialize optimizers
\State Set hyper-parameters and simulation parameters
\State Initialize replay buffer
\State Set training iterations and mini-batches
\State Set PPO-specific parameters

\For{$episode = 1$ \textbf{to} $2,000$}
    \State \textbf{Initialize} episode
    \For{$t = 1$ \textbf{to} $1,000$}
        \State Sample and execute actions
        \State Store transition in the replay buffer
    \EndFor
    \State Update the policy using PPO
\EndFor

\EndProcedure
\label{ag:ppo}
\end{algorithmic}
\end{algorithm}

\subsection*{Autonomous Learning Approach}
To autonomously learn in-hand manipulation of a ball against gravity through utilizing tactile information, we used a model-free RL algorithm to learn the policy. 
We used Proximal Policy Optimization (PPO) as our main algorithm as it presented a balance between the ease of implementation, sample complexity, and ease of adjustment, trying to update at each step to minimize the cost function while assuring that the new policies are not too far from last policies \cite{schulman2017proximal,wang2020reinforcement}. 
PPO has also been adopted as one of the default methods of OpenAI owing to its excellent performance \cite{tang2020implementing,baselines}.

\subsubsection*{\textit{Reward Function}}
The reward engineering concept (a subset of RL) focuses on finding the most appropriate reward to maximize successful learning via reward shaping \cite{mills2020finding}. 
Reward shaping involves carefully designing reward functions that provide the agent with rewards for progress toward the goal. 

In our work, we defined two goals, lift and rotation.
\textbf{Lift:} Our desired height (center of the ball above the ground) is $z_d=25$ mm, shown in (Fig. \ref{fig: overview}).
In our algorithm, the goal is reached when the agent supports the ball against gravity within a desired height range of ~$[21,29]$ mm, indicated with a green box in Figs. \ref{fig: overall_jointplot} (and Supplementary Figs. \ref{fig: O2_jointplots}, \ref{fig: O3_jointplots}, \ref{fig: O4_jointplots}).
A range is used to accommodate height variation during rotation and manipulation tasks.
For results metrics, we report the mean height of the ball and lift success as a percent time within an episode where the ball is in the desired height range. 
\textbf{Rotation:} For rotation, we calculated completed rotations as our performance measurement (as opposed to rotation reward or rotation in degrees).
% We chose the dynamical manipulation task of rotating about a horizontal rotation axis and lifting at the desired height (and adding a penalty proportional to the distance between the current height and the desired height). 
Since we care about manipulating the ball against gravity at the desired height range, we used a combination of primary (positive) reward and punishment (penalty proportional to the distance between the current height and the desired height as a negative reward) at every time step.

In our reward function, the angular velocity of the ball~$\dot{\theta}_{y}$ was the primary reward, and the absolute distance of the state from the reference state of having the ball at the fixed desired position ($z_d=25$ mm, (Fig. \ref{fig: overview})) was the punishment. 
The reward function is described by 

\begin{ceqn}
 \begin{equation}\label{eq:reward}
    Reward_t=c_R\dot{\theta}_{y,t} - c_L|z_{h,t}-z_d|,
\end{equation}   
\end{ceqn}

\noindent where $c_R=0.51$ and $c_L=0.49$. 

We investigated learning strategies (here, curriculum) in which lift and rotation are both rewarded (L+R), strategies in which only lift is rewarded with rotation coefficient set to zero ($c_R$), and strategies in which lift coefficient is set to zero (($c_L$) and only rotation is rewarded (R).
This is described in detail in the following section (see Table \ref{tab:curricula}). 
%%% Move to PPO Learning part if not already said there %%%
% As mentioned previously, not all of the simulated state variables are used in the PPO policy. Although the ball's state is used for the reward function, it is not passed into the PPO policy for learning. In other words, the dynamic interaction with the ball happens in the absence of direct vision (or any other direct, constant monitoring means for the balls velocity or position).
\subsubsection*{\textit{Curriculum Learning}}
A learning trial consisted of 2,000 episodes, where each episode lasted 10 seconds. This resulted in a total simulated time of 5 hours and 33 minutes per trial. Each learning trial was split into two equal halves where the reward function changed between the two halves of the trial. 
We considered five distinct curricula that differed in the behavior (rotation and lift) rewarded in two halves of the trail.
This is illustrated by a circle with a curved arrow (rotation) and a vertical arrow (lift) throughout the paper and pictured in the second column of Table \ref{tab:curricula}). As shown in the last column, by changing $c_R$ and $c_L$ variables in equation \eqref{eq:reward}, we update the reward function in two equal halves of the learning trial in each curriculum.
The final column of the table, gives the values of $c_R$ and $c_L$ used in equation \eqref{eq:reward}, to update the reward function in the two halves of the learning trial in each curriculum.

Learning was evaluated over 60 trials for each of the 5 curricula. 
Each of these 60 trials was independent by varying the seed parameters of the PPO algorithm for our reinforcement learning policy.
This was repeated for the two tactile conditions (No-tactile and 3D-force).
For each tactile condition, the initial seed for the random number generator was held constant across different curricula. 
For example, the first trail run seed was exactly the same for all curricula and both tactile conditions. 
Overall, we used independent trials to evaluate the effectiveness of our approach to autonomous manipulation. 

\begin{table}[ht]
\begin{center}
{\scriptsize
\begin{tabular}{ c c c c c} 
\hline \hline
\vtop{ } & \vtop{\hbox{\strut Reward During First}\hbox{\strut Half of the Trial}} & \vtop{\hbox{\strut Reward During Second }\hbox{\strut Half of the Trial}} 
& 
\vtop{\hbox{\strut Coefficients of equation (\ref{eq:reward}) }\hbox{\strut [first | second] halves of each trial}}\\ [1em]

\hline
 Curriculum 1& 
 \rule{0pt}{5ex}
\raisebox{-.5\totalheight}{
% \rule{0pt}{4ex} 
\includegraphics[width=0.04\textwidth, height=7.2mm]{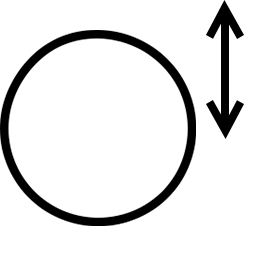}}
 & \raisebox{-.5\totalheight}{ \includegraphics[width=0.1\textwidth, height=7.2mm]{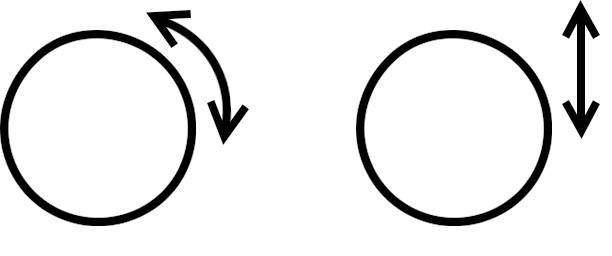}}  & 
 \vtop{\hbox{\strut [L | R+L] }\hbox{\strut  [$c_R=0$, $c_L=0.49$ | $c_R=0.51$, $c_L=0.49$] }}

\\[2em]

\hline 
 Curriculum 2 & 
\rule{0pt}{5ex}
\raisebox{-.5\totalheight}{
% \rule{0pt}{4ex} 

\includegraphics[width=0.04\textwidth, height=7.2mm]{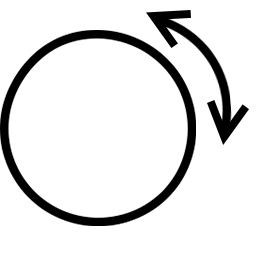}}
 & \raisebox{-.5\totalheight}{ \includegraphics[width=0.1\textwidth, height=7.2mm]{Figures/Table_ballLR.jpg}} & \vtop{\hbox{\strut [R | R+L] }\hbox{\strut  [$c_R=0.51$, $c_L=0$ | $c_R=0.51$, $c_L=0.49$] }} \\[2em]

\hline
Curriculum 3 & 
\rule{0pt}{5ex}
\raisebox{-.5\totalheight}{\includegraphics[width=0.1\textwidth, height=7.2mm]{Figures/Table_ballLR.jpg}} & \raisebox{-.5\totalheight}{\includegraphics[width=0.1\textwidth, height=7.2mm]{Figures/Table_ballLR.jpg}} & \vtop{\hbox{\strut \quad [R+L | R+L] }\hbox{\strut \quad [$c_R=0.51$, $c_L=0.49$ | $c_R=0.51$, $c_L=0.49$] }}
\\[2em] 

\hline
Curriculum 4 & 
\rule{0pt}{5ex}
\raisebox{-.5\totalheight}{ \includegraphics[width=0.1\textwidth, height=7mm]{Figures/Table_ballLR.jpg}}
 & \raisebox{-.5\totalheight}{ \includegraphics[width=0.04\textwidth, height=7.2mm]{Figures/Table_ballR.jpg}}  & 
 \vtop{\hbox{\strut [R+L | R] }\hbox{\strut  [$c_R=0.51$, $c_L=0.49$ | $c_R=0.51$, $c_L=0$] }}
\\[2em]

\hline
Curriculum 5 &
\rule{0pt}{5ex}
\raisebox{-.5\totalheight}{\includegraphics[width=0.1\textwidth, height=7mm]{Figures/Table_ballLR.jpg}}& \raisebox{-.5\totalheight}{\includegraphics[width=0.04\textwidth, height=7.2mm]{Figures/Table_ballL.jpg}}  &
 \vtop{\hbox{\strut [R+L | L] }\hbox{\strut  [$c_R=0.51$, $c_L=0.49$ | $c_R=0$, $c_L=0.49$] }}
\\[2em] 

\hline \hline
\end{tabular}
}
\end{center}
\captionsetup{singlelinecheck=off,font=small}
 \caption{We used five curricula that rewarded different combinations of rotation and lift during each half of the independent trials.
These changes in the coefficients of the reward function define a progression of goals (i.e., curriculum learning) over the two halves of each run.}
\label{tab:curricula}
\end{table}

\subsubsection*{\textit{Reinforcement Learning Policy}}

We adopted a Proximal Policy Optimization (PPO) policy to control the robotic hand to achieve autonomous manipulation. 
PPO is a set of policy gradient methods that optimize a surrogate objective function using multiple minibatch updates per data sample \cite{schulman2017proximal, kristensen2020strategies}. The objective function to optimize is the sum of several loss functions and is given by
\begin{ceqn}
\begin{equation}\label{eq:ppo}
	L_t^{
		CLIP +V F +S}
	(\theta) = \hat{\mathbb{E}}_t[L_{t}^{CLIP}(\theta)-c_1 L^{VF}_t(\theta) + c_2S[\pi_{\theta}](s_t)],
\end{equation} 
\end{ceqn}

\noindent The~$L_{t}^{CLIP}(\theta)$ is the clipped surrogate loss function and ensures that the policy updates will not be too large. 
While the~$L^{VF}_t$ is a squared-error loss, it ensures that the loss from both policy and value functions of the neural networks are accounted for. The~$S$ denotes the entropy bonus term, which encourages a more random policy (i.e., more exploration), so a larger entropy coefficient~$c_2$ will encourage more exploration  \cite{kristensen2020strategies}.

To implement PPO, we use the PPO1 implementation from OpenAI's stable baselines repository \cite{baselines} with MultiLayer Perceptron (MLP) Artificial Neural Network (ANN) for the actor-critic mapping. The Proximal Policy Optimization (PPO) algorithm, outlined in Algorithm \ref{ag:ppo}, describes the iterative process through which the policy and value functions are updated to maximize cumulative rewards.

At every time step~$t$, the robotic hand observes its own state ~$\mathbf{s}_{h,t}$ and the state of the ball~$\mathbf{s}_{b,t}$, predicts the optimized action, executes it~$\mathbf{a}_t$, and a reward is used~$r_t$.
The state~$\mathbf{s}_{h,t}$ contains the angle and angular velocity~$q_t,\dot{q}_t $ of each finger and the position and linear velocity of the palm at every time step~$t$. The overview diagram of the Proximal Policy Optimization algorithm in this work is shown in (Fig. \ref{subfig_overview learning}).

To attain optimal performance, we fine-tuned the hyper-parameters and meta-paratmers of the PPO algorithm during the training of our RL model.
The clipped surrogate loss in the PPO algorithm serves to prevent divergence, as discussed in \cite{schulman2017proximal}. However, it introduces a challenge by potentially prematurely reducing exploration variance across multiple iterations. Strategic tuning of the loss parameters in equation \eqref{eq:ppo} becomes essential to avoid issues such as divergence or settling on a local minimum. PPO addresses this challenge by including an entropy loss term that penalizes low variance, mitigating the risk of premature convergence. It has been observed that a higher entropy loss weight minimizes the risk of getting trapped in local optima. Nevertheless, if the entropy loss weight is excessively large, it can lead to a noisy policy and a decline in average performance. Therefore, careful adjustment of the entropy loss term for PPO is necessary. Building on the findings regarding different entropy loss weights for the policy’s standard deviation in \cite{hamalainen2020ppo}, we optimized the entropy loss term to strike a balance between variance and average performance.

PPO employs the Generalized Advantage Estimator (GAE) to reduce the variance of policy gradient estimates at the expense of some tolerable bias.
GAE is parameterized by $\lambda  \in  [0, 1]$, which enables the PPO agent with a mechanism to control policy updates according to the significance of each sampled state and, therefore, enhance learning reliability \cite{chen2018adaptive}. Changing this hyper-parameter enables PPO to find a balance between variance and bias of policy gradient estimates \cite{schulman2015high}. In our work, this trade-off was achieved by changing the lambda meta-parameter to relatively demote rewards achieved later in the episode (when the ball may have been dropped) and instead emphasizing immediate rewards at every point in time (as is the case in real life).

The number of optimization epochs, GAE parameter~$\lambda$, and the entropy coefficient are set to values shown in Table \ref{tab:ppo}. All other parameters are kept at their default values per PPO implementation.

\subsubsection*{\textit{Adaptive curriculum-based learning rate scheduler}}
The impact of learning rate scheduling on the performance of stochastic optimizers has garnered considerable attention in recent research \cite{cooper2021hyperparameter, xu2019learning}. Traditional approaches, employing a fixed and static learning rate throughout training, often struggle to attain optimal model performance.
To address this limitation, diverse scheduling algorithms, such as polynomial decay, cosine decay, and warm-up, have been proposed, each tailored with distinctive forms \cite{gotmare2018closer}. 
Current methodologies often rely on predefined principles, assuming specific scheduling rules based on empirical studies and domain knowledge. These approaches may not rigidly adhere to any existing rule to find the optimal learning rate scheduling for a particular problem.

In our exploration of PPO, we aim to transcend the constraints associated with a constant learning rate.
Initially, we opted for a constant learning rate and linear learning rate, commonly used approaches in reinforcement learning algorithms \cite{xiong2022learning}. 
But implementing a constant learning rate in dynamic contexts, where sensitivity to the initial rate choice can result in unstable training or sub-optimal solutions, highlights the necessity for adaptive approaches.
We proposed a new method to tackle challenges with fixed learning rates, especially in dynamic environments like our manipulation tasks. 
This is addressed by an adaptive curriculum-based learning rate scheduler, bringing multiple advantages.
This adaptive strategy dynamically adjusts rates throughout training, expediting the learning process with higher initial rates and ensuring stable convergence through decrementing rates during later stages.

\subsubsection*{Curriculum-based Learning Rate Scheduler Strategy}

Instead of utilizing a fixed or decreasing learning rate, our method embraces a curriculum-adaptive learning strategy.
The adaptive curriculum-based learning rate scheduler (piecewise linear learning
rate) strategy is described as follows:
\begin{equation}\label{eq:learning}
Lr=\begin{cases} 
   \phi \cdot \left(1 - \frac{\text{{sample number}}}{1,000,000}\right), & \text{sample number} \leq 1,000,000 \\
    \eta \cdot \left(1 - \frac{\text{{sample number}}}{2,000,000}\right), &  \text{sample number} > 1,000,000
\end{cases}    
\end{equation}

The selection of optimal values for \( \phi\) and \( \eta\) was determined empirically, ensuring adaptability across all five curricula.
The curriculum-based learning rate scheduler (\(Lr\)) is established and adjusted through trial and error to emphasize the significance of curriculum learning. These coefficients are then integrated into the PPO linear scheduler according to the following equation.
Our curriculum dynamically changes at 1,000 episodes (1,000,000 samples), compelling the learning rate to be piecewise linear to accommodate the variations in the dynamics of the reward and tasks. This adaptive strategy effectively responds to changes in the environment, contributing to the model's success.

\begin{figure*}[ht!]
  \centering
  \begin{subfigure}[t]{0.95\textwidth}
    \centering
      \includegraphics[width=\textwidth]{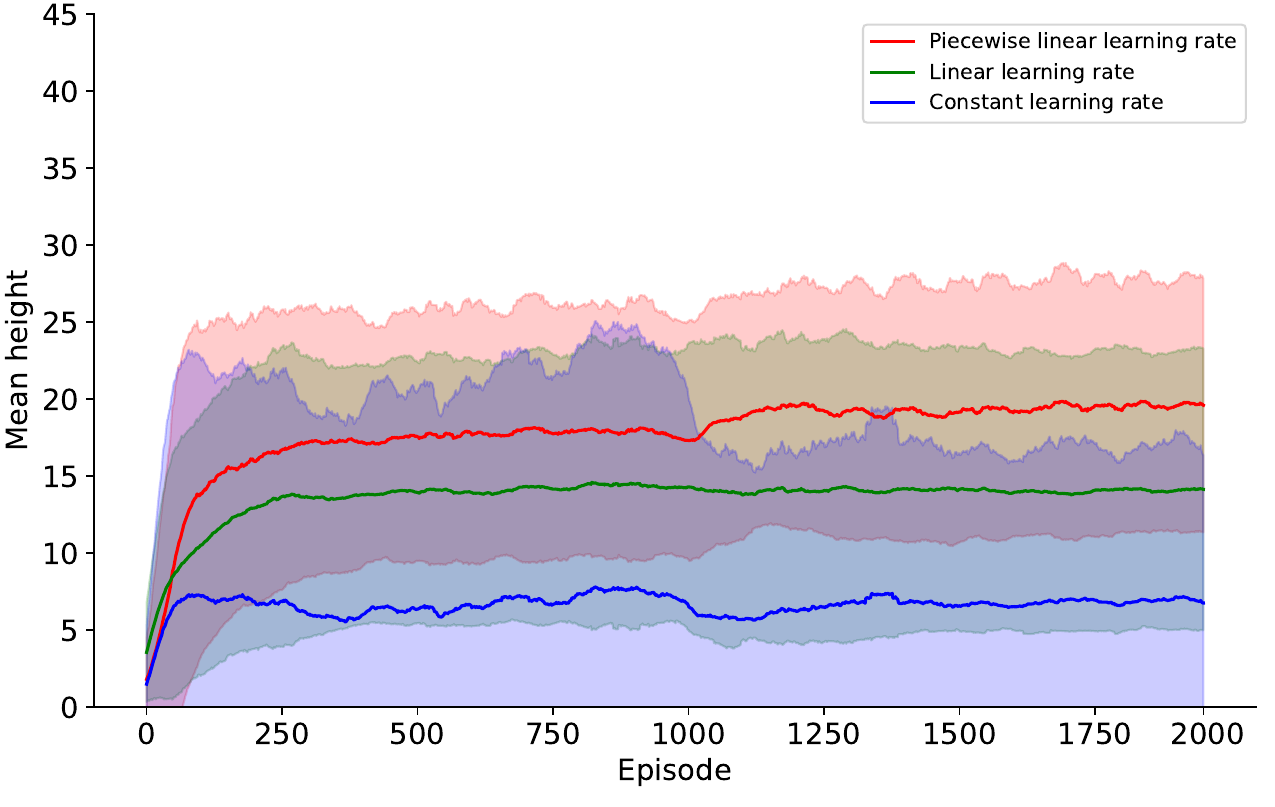}
      % \caption[]%
      {{}}
      \label{subfig_c5_jointplot}
    \end{subfigure}
    % \centerline{\rule{16cm}{1pt}}
     \caption[]
    {\footnotesize \textbf{
    Effect of PPO curriculum-based learning rate scheduler by comparison of Mean Height.} 
    Data presented for mean height in \Cfive throughout the whole learning period. The desired height for all cases is $25$ mm. Solid lines represent the mean across all 60 trials for the specified learning rate methods. Shaded areas represent $\pm 1$ standard deviation.
    The solid red line follows the PPO implementation per equation 
    \eqref{eq:learning}.
   }
    \label{fig:learningrate}
\end{figure*} 

To validate the effectiveness of our approach, we explore the impact of constant, linear, and adaptive curriculum-based learning rate scheduler (Piecewise Linear Learning
Rate) in \Cfive in Fig. \ref{fig:learningrate}, comparing Mean height over 2,000 learning episodes. 
% Desired height for all cases was 25mm. 
Piecewise linear learning rate was far closer to this target height than either Linear or Constant learning rate.
Thus piecewise linear learning rate was used as the curriculum-based learning rate scheduler throughout this work.
Our results in different curricula consistently support the superior performance of PPO with well-designed scheduling mechanisms, surpassing those utilizing a constant and linear learning rate in both convergence rate and final performance metrics \cite{senior2013empirical, gotmare2018closer,loshchilov2016sgdr}.
One key advantage is the reduced sensitivity to the initial rate choice, minimizing the risk of divergence.
The piecewise linear learning rate promotes efficient exploration in the early stages and exploitation for optimal performance during convergence. 
Its curriculum-based adaptive nature contributes to faster convergence, effectively navigating both exploratory and exploitative learning phases.
Moreover, the piecewise linear schedule imparts robustness against variations in task difficulty or environmental changes, automatically adjusting to maintain training stability.

To evaluate the effectiveness of different learning scheduler methods in reducing convergence time, we conducted an analysis on the average number of episodes needed after switching the reward during the second phase of learning in \Cfive. We compared three learning schedulers: constant learning rate, linear rate, and piecewise linear rate.

Our findings reveal that the average number of episodes for convergence in successful trials (defined as trials where the hand can maintain the ball within the target height range) after the reward switch varied significantly across the different schedulers.
Specifically, when focusing only on the successful trials (not shown), we observed that it took 1,000 episodes for convergence with a constant learning rate, 450 episodes with a linear rate, and only 250 episodes with a piecewise linear rate (see episodes 1,250 in Fig. \ref{fig:learningrate}).

Figure \ref{fig:learningrate} illustrates the performance of each scheduler in reaching the target height. Remarkably, the piecewise linear learning rate outperformed both the linear and constant rates by a substantial margin. Additionally, it achieved a higher cumulative reward across all 60 trials, indicating its superior effectiveness in learning and adaptation.

These results highlight the significant advantages of using a piecewise linear learning rate scheduler in enhancing convergence speed and overall performance in \Cfive simulations.
% \begin{table}[]
% \centering 
% \begin{tabular}{p{6 cm} p{7 cm}}
% \hline\hline
% \textbf{Learning Scheduler} & \textbf{Average Number of Episodes for Convergence in Successful Trials After the Switch} \\ \hline 
% Constant Learning Rate & 1000 \\ %\hline
% Linear Learning Rate & 450 \\ %\hline
% Piecewise Linear Learning Rate & 250 \\ \hline
% \hline\hline
% \end{tabular}
% \caption{Impact of learning schedulers on PPO learning speed. The piecewise linear learning rate is a curriculum-adaptive learning strategy explained in \eqref{eq:learning}.}
% \label{tab:ppo_learning_schedulers}
% \end{table}

In summary, our adaptive curriculum-based learning rate scheduler strategy in the PPO implementation aims to enhance training stability, expedite convergence, and improve adaptability in dynamic environments. This aligns with our goal of efficiently training the agent for effective in-hand manipulation and contributes to the exploration of learning rate scheduling strategies on a curriculum-based approach. 

The complete code for learning is available at the following \href{https://github.com/pojaghi/In-hand-manipulation} {GitHub repository}.

\bibliographystyle{unsrt} 

\bibliography{scifile}

\section*{Acknowledgements}
This work is supported in part by the NIH R21 NS113613-01A1, NSF 2113096 CRCNS US-Japan, DOD CDMRP Grant MR150091, DARPA-L2M W911NF1820264 awarded to FV-C, and the USC Viterbi School of Engineering Fellowships to AM and RM. This work does not necessarily represent the views of the NIH, NSF, DoD, or DARPA.

This work was supported in part by the University of Wisconsin-Madison Office of the Vice Chancellor for Research and Graduate Education funded by the Wisconsin Alumni Research Foundation.

\section*{Author contributions}
P.O., R.M., A.M., M.W., and F.J.V.-C. contributed to the conception and design of the work. 
P.O., R.M., A.M., and A.E. contributed to the data analysis.
F.J.V.-C., M.W., and A.M. provided general direction for the project. 
All authors approved the final version of the manuscript and agree to be accountable for all aspects of the work. 
All persons designated as authors qualify for authorship, and all those who qualify for authorship are listed.

%\section*{Additional information}

\section*{Competing Interests}
The authors declare no competing interests.

\newpage
\section*{Supplementary Information}
\renewcommand{\thefigure}{S\arabic{figure}}
\setcounter{figure}{0}
\renewcommand{\thetable}{S\arabic{table}}
\setcounter{table}{0}

\subsection*{Supplementary Methods}  

We simulate a three-fingered robotic hand using MuJoCo, an advanced physics simulation environment that enables us to model physical systems and their dynamic interactions. We also utilize the OpenAI baselines library and the MuJoCo-py interface to implement reinforcement learning algorithms developed in Python. The robotic hand attempts to autonomously learn to manipulate a ball (Fig. \ref{subfig:overview simulation}). Our data-driven learning approach does not necessitate an explicit model of the hand, ball, or their interactions, which is advantageous in unstructured environments compared to model-based approaches \cite{cruciani2020benchmarking,kumar2014real}. Furthermore, our approach does not rely on visual information at any stage of the learning process.

Our aim is to achieve autonomous acquisition of manipulation skills by the robotic hand, with upcoming sections detailing the simulation, learning algorithm, and thoroughly exploring the assessment of generalizability and robustness.

\subsubsection*{Generalizability}

To showcase the versatility and resilience of our proposed methodologies, we conduct thorough evaluations across four distinct objects. These evaluations involve systematic exploration, incorporating various weight combinations such as 5g and 50g weights, along with different ball radii (35 mm and 30 mm). Physical simulation parameters for the three-fingered robotic hand and parameters of the ball for different objects are given in Tables \ref{tab: hand_parameter} and \ref{tab: simulation parameters generalizability}.

Within the scope of our investigation, we carry out a series of exhaustive experiments for each of the four distinct objects, denoted as O1 through O4. These experiments are carefully designed, incorporating five diverse curricula identified as C1 through C5. A rigorous evaluation, comprising a total of 60 trials for each curriculum, provides valuable insights into the adaptability and performance of our methods. In addition, we systematically explored two tactile conditions (No-tactile vs. 3D-force) across the five distinct curricula.
The subsequent section provides detailed simulation parameters for all four distinct objects.

\subsubsection*{Simulation parameters}

Physical parameters for all entities in the simulation must be specified (either directly or indirectly) including size, mass, stiffness, and damping. Relevant simulation parameters for the hand and ball for four objects are provided in Table \ref{tab: hand_parameter} and \ref{tab: simulation parameters generalizability}.

\begin{table}[]  
\centering % centering table  
\begin{tabular}{l c r} 
\hline\hline   
 Parameter &\multicolumn{1}{c}{Value}  
\\ [0.5ex]  
\hline   
% Hand
Palm Mass
& 100\ g \\[-0.2ex] 
% & 76.02\ gram \\[-0.2ex] 
Finger Mass  
& 68 g  \\[0 ex] 
Link length 
 & 50 mm \\[-0.2ex]  
Phalanx diameter  
& 10 mm \\[-0.2 ex]
Palm width
& 20 mm  \\ [-0.2 ex]
Palm diameter  
& 120 mm  \\[-0.2 ex]
Initial hand height~($z_h(0)$)  & 200 mm  \\[0.2ex]
Maximum translation~($z_h$)  &
130 mm  \\[0.2ex]
Joint damping &
$5.5 \times 10^{-6}$ N$\cdot$s/mm\\[0.2ex] 
Joint limits ($q_1$)
&~$[-45^{\circ},45^{\circ}]$\\[-0.2 ex]
Joint limits ($q_2$)
&~$[-90^{\circ},0^{\circ}]$\\[2 ex]
\hline \hline % inserts single-line  
\end{tabular} 
\caption{{Physical simulation parameters for the three-fingered robotic hand for all objects. 
The overall mass of the hand comprises the combined masses of the three fingers and the palm is 304 g.} 
} 
\label{tab: hand_parameter} 
\end{table}

\begin{table}[]  
\centering % centering table  
\begin{tabular}{l c c r} 
\hline\hline   
Object & Parameter &\multicolumn{1}{c}{Value}  
\\ [0.5ex]  
\hline   
% Ball 
 &Mass  & 50 g \\[-0.2ex]  
\raisebox{1.5ex}{} & Radius   
& 35 mm  \\[0 ex] 
\raisebox{1.5ex}{} &Desired height~($z_d$)
& 60 mm \\[-0.2 ex]
\raisebox{1.5ex}{} &Height~($z_b$)
& 35 mm \\[0ex]  
\raisebox{1.5ex}{} & Stiffness in ~$x$ direction   
& $5 \times 10^{-3}$ N/mm\\[-1 ex]
\raisebox{1.5ex}{Object 1} & Damping in ~$x$ direction 
& $3.5 \times 10^{-4}$ N$\cdot$s/mm \\ [-0.5 ex]
\raisebox{1.5ex}{} &Damping in ~$z$ direction  
& $5 \times 10^{-4}$ N$\cdot$s/mm \\[-0.2 ex]
\raisebox{1.5ex}{} &Damping about ~$y$ direction  
& $5 \times 10^{-3}$ N$\cdot$s/rad \\[1 ex]

% Ball 
 &Mass  & 50 g \\[-0.2ex]  
\raisebox{1.5ex}{} & Radius   
& 30 mm  \\[0 ex] 
\raisebox{1.5ex}{} &Desired height~($z_d$)
& 60 mm \\[-0.2 ex]
\raisebox{1.5ex}{} &Height~($z_b$)
&30 mm \\[0ex]  
\raisebox{1.5ex}{} & Stiffness in ~$x$ direction   
& $5 \times 10^{-3}$ N/mm\\[-1 ex]
\raisebox{1.5ex}{Object 2} & Damping in ~$x$ direction 
& $3.5 \times 10^{-4}$ N$\cdot$s/mm \\ [-0.5 ex]
\raisebox{1.5ex}{} &Damping in ~$z$ direction  
& $2 \times 10^{-4}$ N$\cdot$s/mm \\[-0.2 ex]
\raisebox{1.5ex}{} &Damping about ~$y$ direction  
& $5 \times 10^{-3}$ N$\cdot$s/rad \\[1 ex]

% Ball 
 &Mass  & 5 g \\[-0.2ex]  
\raisebox{1.5ex}{} & Radius   
& 35 mm  \\[0 ex] 
\raisebox{1.5ex}{} &Desired height~($z_d$)
& 60 mm \\[-0.2 ex]
\raisebox{1.5ex}{} &Height~($z_b$)
&35 mm \\[0ex]  
\raisebox{1.5ex}{} & Stiffness in ~$x$ direction   
& $1 \times 10^{-3}$ N/mm\\[-1 ex]
\raisebox{1.5ex}{Object 3} & Damping in ~$x$ direction 
& $7 \times 10^{-5}$ N$\cdot$s/mm \\ [-0.5 ex]
\raisebox{1.5ex}{} &Damping in ~$z$ direction  
& $2 \times 10^{-4}$ N$\cdot$s/mm \\[-0.2 ex]
\raisebox{1.5ex}{} &Damping about ~$y$ direction  
& $1 \times 10^{-3}$ N$\cdot$s/rad \\[1 ex]

% Ball 
 &Mass  & 5 g \\[-0.2ex]  
\raisebox{1.5ex}{} & Radius   
& 30 mm  \\[0 ex] 
\raisebox{1.5ex}{} &Desired height~($z_d$)
& 60 mm \\[-0.2 ex]
\raisebox{1.5ex}{} &Height~($z_b$)
&30 mm \\[0ex]  
\raisebox{1.5ex}{} & Stiffness in ~$x$ direction   
& $1 \times 10^{-3}$ N/mm\\[-1 ex]
\raisebox{1.5ex}{Object 4} & Damping in ~$x$ direction 
& $7 \times 10^{-5}$ N$\cdot$s/mm \\ [-0.5 ex]
\raisebox{1.5ex}{} &Damping in ~$z$ direction  
& $2 \times 10^{-4}$ N$\cdot$s/mm \\[-0.2 ex]
\raisebox{1.5ex}{} &Damping about ~$y$ direction  
& $1 \times 10^{-3}$ N$\cdot$s/rad \\[1 ex]

\hline \hline  
\end{tabular} 
\caption{ {Simulation parameters for the ball across all objects, detailing size, mass, stiffness, and damping.} 
} 
\label{tab: simulation parameters generalizability} 
\end{table}

\begin{figure*}[t!]
\centerline{\includegraphics[width=.8
\linewidth]{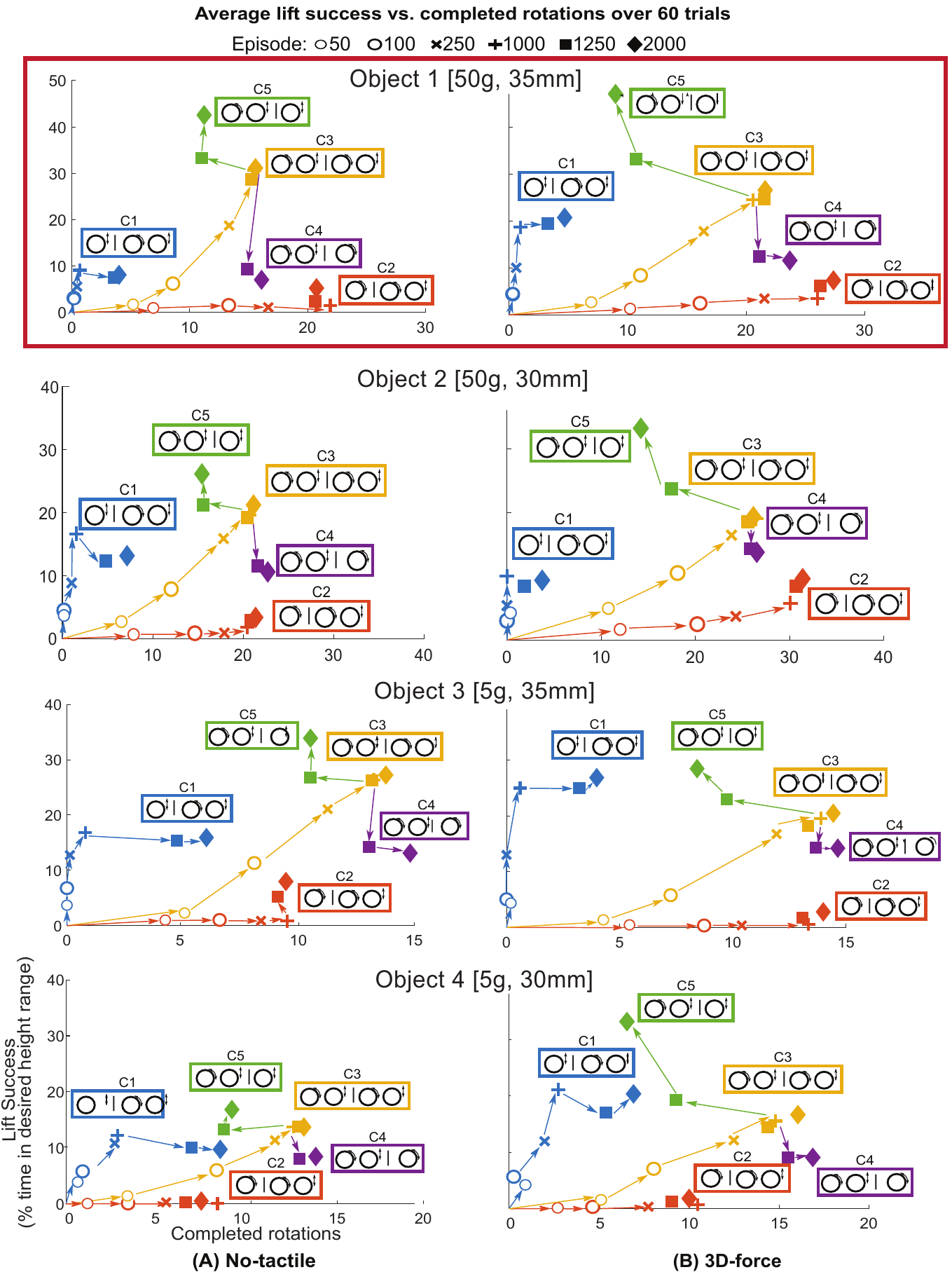}}
    \caption[]
    {\footnotesize \textbf{The evolution of learning highlights the dynamic functional interaction between curriculum and tactile information across all objects. }
    Manipulation performance during the last 10 seconds of each episode is highlighted. The graph depicts lift success (percentage of time the ball remains within the desired height range) versus completed rotations. Each data point represents the average of 60 independent trials, with arrows indicating the direction of increasing episodes. Negative rotations were constrained to zero. It's worth noting the significant impact of curriculum selection on learning outcomes, observed across both tactile conditions (\textbf{(A)} No-tactile and \textbf{(B)} 3D-force). The Red box is the default object. 
   }
    \label{fig: all_overview}
\end{figure*}

\subsubsection*{Tactile information.}
Tactile information is provided to the learning policy via the tactile force state vector for the simulated robotic hand ($\mathbf{s_{h,f}}$).
Contact force is measured at the pad of the fingers (where the tactile area is defined, see (Fig. \ref{subfig:overview simulation})), and directed from the finger towards the ball  \cite{todorov2012mujoco}, which at most contains the full touch force vector $\mathbf{f}= [f_{t1}, f_{t2}, f_n]$. Each finger-pad outputs the tactile information independent from other pads. The learning policy has two options for tactile information, which are added to the state vector for the hand ($\mathbf{s_h}$). These are No-tactile or 3D-force information. Table \ref{tab:tactile force info} indicates the detailed tactile information for the two tactile options.

\begin{table}[ht]
\centering
 \begin{tabular} { p {2 cm}  p {4.5 cm} p {2.5 cm} p {3.5 cm}}
 \hline \hline
 \multicolumn{4} { c }{Tactile Information in State Variable}\\
 \hline
 No-tactile &3D-force\\
 \hline
 $s_{h,f}=0$ & $s_{h,f}=[f_{t1},f_{t2},f_n]$ \\
 \hline \hline
 \end{tabular}
 \caption{ {Tactile information options available to the learning policy.} 
 }
 \label{tab:tactile force info}
\end{table}

\subsubsection*{Leaning manipulation through PPO}
Proximal Policy Optimization (PPO) comprises a collection of policy gradient methods designed to optimize a surrogate objective function through multiple minibatch updates per data sample \cite{schulman2017proximal, kristensen2020strategies}. 
PPO leverages the Actor-Critic Model, which consists of two Deep Neural Networks. One network is responsible for action selection (the actor), while the other handles reward estimation (the critic).
We conducted a series of experiments to assess the performance of the PPO algorithm in our environment. The following section provides an in-depth examination of the hyperparameters for PPO, which were determined through trial and error, and a comprehensive analysis of the resulting performances.

% The objective function to be optimized is a composite of various loss functions and can be expressed as:

% \begin{ceqn}
% \begin{equation}\label{eq:ppo_detail}
% L_t^{CLIP + VF + S}(\theta) = \hat{\mathbb{E}}t[L{t}^{CLIP}(\theta) - c_1 L^{VF}_t(\theta) + c_2S\pi_{\theta}],
% \end{equation}
% \end{ceqn}

% Here, $L_{t}^{CLIP}(\theta)$ serves as the surrogate objective function, ensuring that policy updates remain within reasonable bounds. The term $L^{VF}_t$ corresponds to a squared-error loss, accounting for losses from both the policy and value functions within the neural networks. The symbol $S$ represents the entropy bonus term, encouraging a more exploratory policy, thereby promoting greater randomness in decision-making. Consequently, a larger entropy coefficient, denoted as $c_2$, encourages increased exploration \cite{kristensen2020strategies}.

\textit{\underline{{PPO Hyper-parameter}}}

The robotic agent learns manipulation using OpenAI Baselines' PPO1 implementation as the RL algorithm. We meticulously select hyperparameters to emphasize rewards at each simulation time step over those at the end of episodes. Additionally, we employ a learning scheduler customized for the established proximal-policy optimization (PPO) algorithm \cite{schulman2017proximal}.
The hyper-parameters for the PPO algorithm are listed in Table \ref{tab:ppo}. Non-default hyper-parameters are chosen empirically through trial and error and careful examination of resulting performances.

\begin{table}[ht]
\centering
 \begin{tabular} { p {6 cm}  p {1.4 cm}}
  \hline
  \hline
 hyper-parameter &Value\\
 \hline
 Adam stepsize  & $1 \times 10^{-5}$ \\
 Number of epochs  & 8\\
 Discount~($\gamma$) & 0.99 \\
 Entropy coefficient & 0.02\\
 Advantage estimation~($\lambda$)&0.85\\
 Minibatch size& 64\\
 \hline
 \hline
 \end{tabular}
 \caption{
   {Proximal-policy optimization (PPO)  hyper-parameters} 
   }
\label{tab:ppo}
\end{table} 

\subsubsection*{Linear Rate Scheduler}

Employing a fixed and unchanging learning rate throughout the entirety of training often falls short of achieving an optimal model. Recognizing the critical role of learning rate schedules in model performance, researchers have extensively investigated methods for effectively and automatically tuning the learning rate for stochastic optimizers. This is because stochastic optimizers are highly sensitive to the learning rate scheduling \cite{cooper2021hyperparameter, xu2019learning}.

\textit{\underline{{The Constant Learning Rate Dilemma}}}

Implementing a constant learning rate, while straightforward in stationary or slowly changing environments, presents challenges in dynamic contexts. Sensitivity to the initial rate choice is a primary limitation, with high rates risking unstable training and low rates leading to protracted convergence or suboptimal solutions. Achieving the right balance becomes a nuanced trial-and-error process, demanding significant time and effort. Furthermore, the fixed learning rate lacks adaptability to evolving learning dynamics, resulting in suboptimal training efficiency and performance.

\textit{\underline{{The Appeal of Linearly Changing Learning Rates.}}}

In response to these challenges, we transitioned to a linearly changing learning rate schedule, offering several advantages. This adaptive approach dynamically adjusts rates throughout training, expediting the learning process with higher initial rates and facilitating stable convergence through decrementing rates during the later stages.

One significant advantage is the reduced sensitivity to the initial rate choice, minimizing the risk of divergence. The linearly changing learning rate promotes efficient exploration by encouraging policy discovery in the early stages and exploitation for optimal performance during convergence. Its adaptive nature contributes to faster convergence compared to a fixed rate, effectively navigating both exploratory and exploitative learning phases.

Moreover, the linear schedule imparts robustness against variations in task difficulty or environmental changes, automatically adjusting to maintain training stability. In summary, our transition to a linearly changing learning rate in the PPO implementation aims to enhance training stability, expedite convergence, and improve adaptability in dynamic environments. This strategy aligns with our goal of efficiently training the agent for effective in-hand manipulation and contributes to the exploration of learning rate scheduling strategies in stochastic optimization. Additionally, we present results for Curriculum 3 [L+R | L+R], comparing training dynamics under both linear and constant learning rate scenarios to gain insights into the effectiveness of different strategies in our experimental setup.

\subsection*{Supplementary Results} 
\begin{figure*}[ht!]
    \centering
      \includegraphics[width=\textwidth]{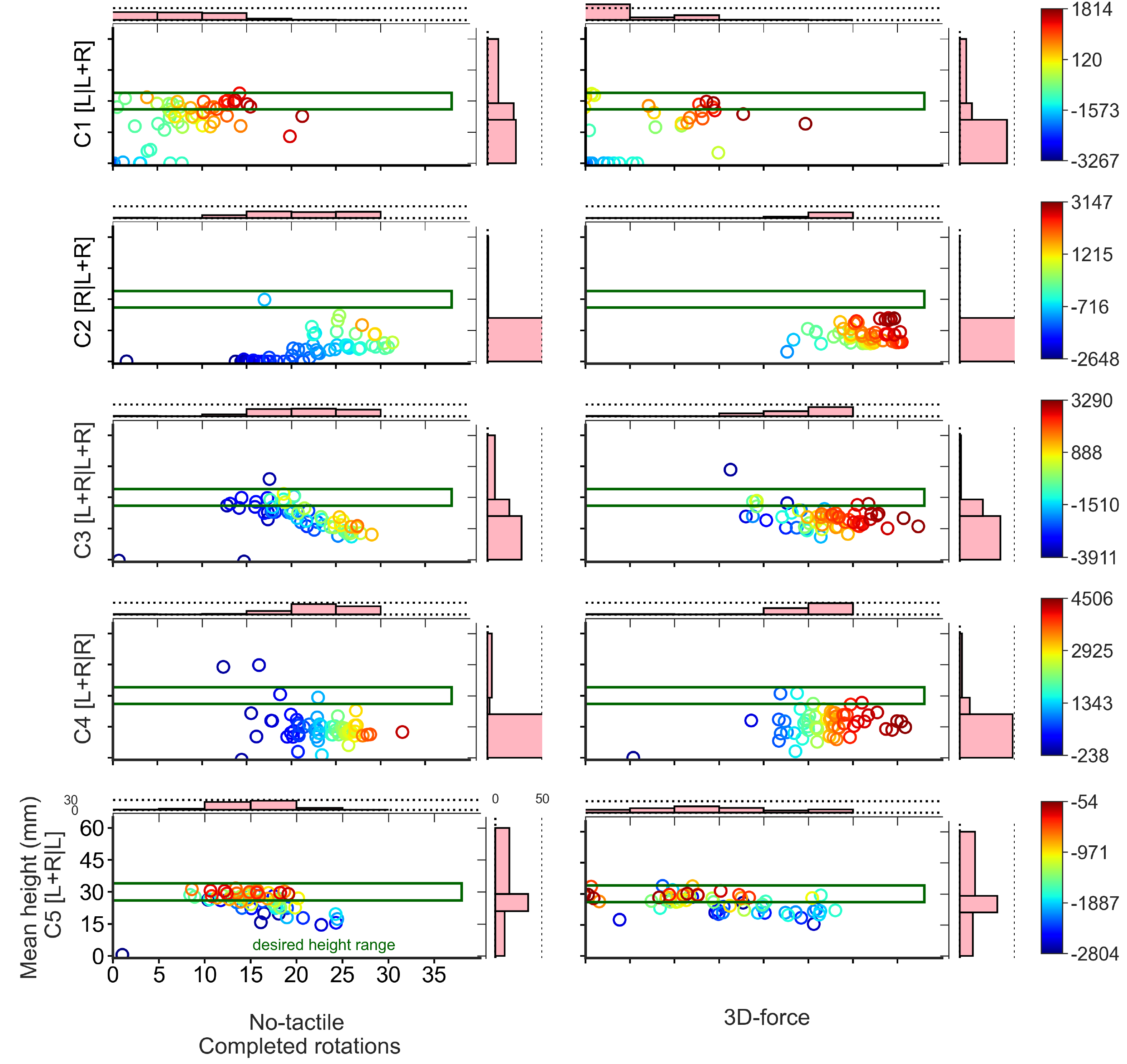}
    \caption{
    {\footnotesize \textbf{Evaluation of performance across all curricula and both tactile information options for Object 2: 50g, 30 mm.}
    The joint distribution illustrates the performance during the final episode of 60 trial runs (showcasing the mean ball height (mm) versus the number of completed rotations). The color-coded cumulative reward for the last episode of each independent run (refer to equation 1) corresponds to different curricula. Note that the desired manipulation performance is represented by those points inside the green box defining the desired ball height (30 ± 4 mm).
    }}
\label{fig: O2_jointplots}
\end{figure*} 

\begin{figure*}[ht]
  \centering
      \includegraphics[width=0.85\textwidth]{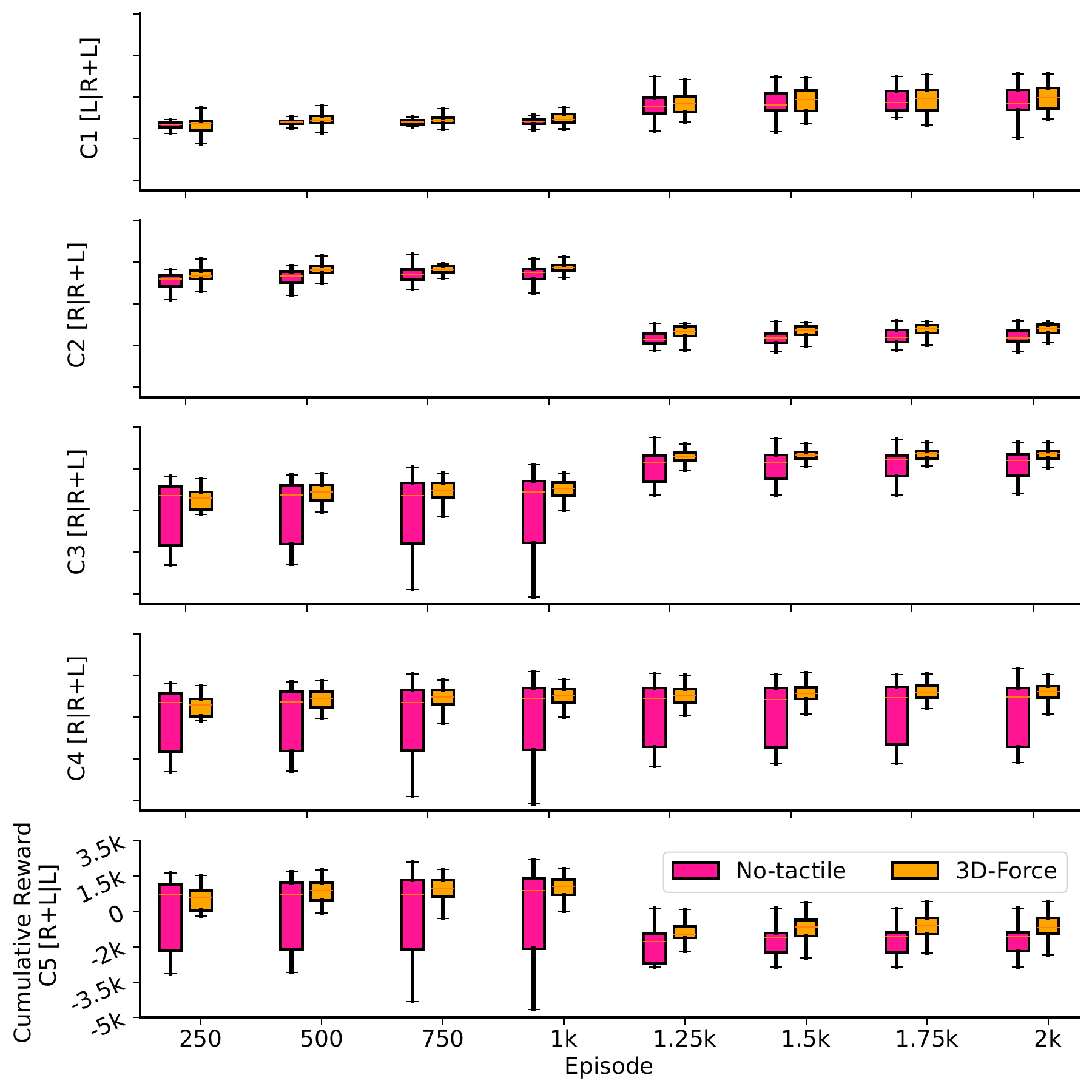}
     \caption{
    {\footnotesize \textbf{Cumulative reward for each representative episode across all curricula and both tactile information options for Object 2: 50 g, 30 mm}. Boxplots, with median, across tactile information options for 60 MC runs at eight representative episodes, 250 episodes apart.
    }}
 \label{fig:S2_boxplots}
\end{figure*}

\begin{figure*}[ht]
  \centering
      \includegraphics[width=\textwidth]{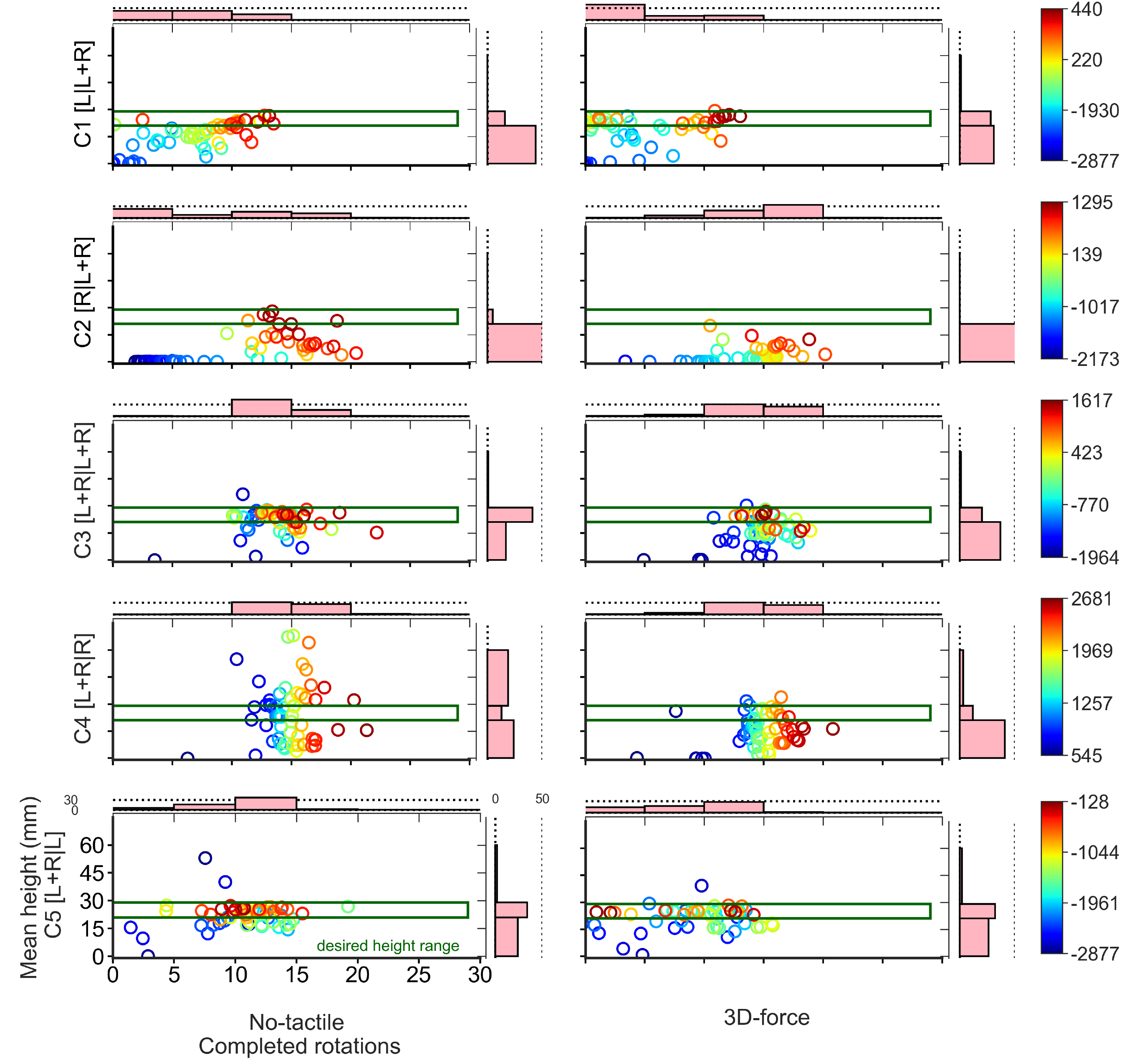}
     \caption
    {\footnotesize \textbf{Evaluation of performance across all curricula and both tactile information options for Object 3: 5 g, 35 mm.}
    The joint distribution illustrates the performance during the final episode of 60 trial runs (showcasing the mean ball height (mm) versus the number of completed rotations). The color-coded cumulative reward for the last episode of each independent run (refer to equation 1) corresponds to different curricula. Note that the desired manipulation performance is represented by those points inside the green box defining the desired ball height (25 ± 4 mm).
    }
 \label{fig: O3_jointplots}
\end{figure*}

\begin{figure*}[ht]
    \centering
      \includegraphics[width=0.85\textwidth]{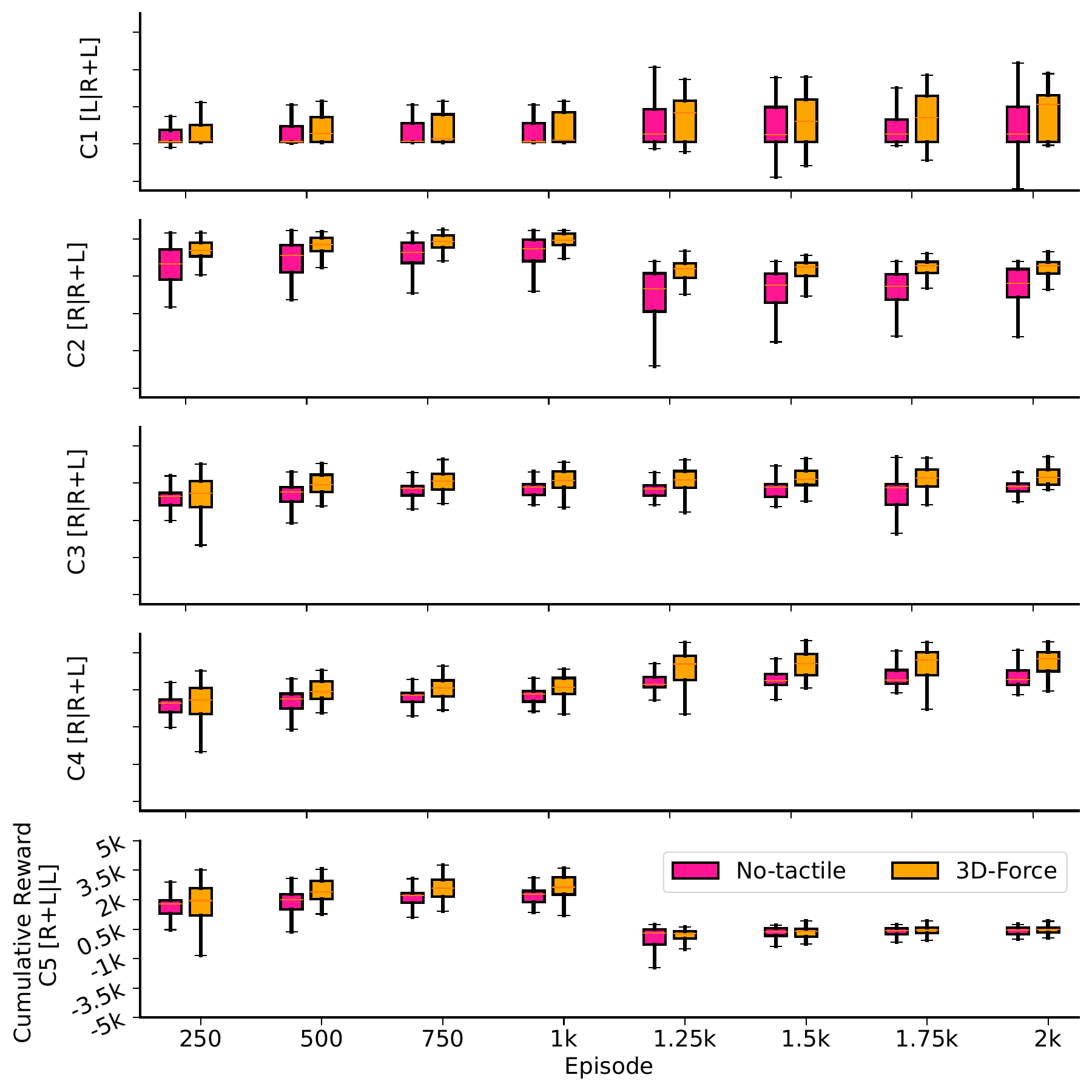}
     \caption[]
    {\footnotesize \textbf{Cumulative reward for each representative episode across all curricula and both tactile information options for Object 3: 5 g, 35 mm}. Boxplots, with median, across tactile information options for 60 MC runs at eight representative episodes, 250 episodes apart.
    }
 \label{fig:S3_boxplots}
\end{figure*} 

\begin{figure*}[ht]
  \centering
      \includegraphics[width=\textwidth]{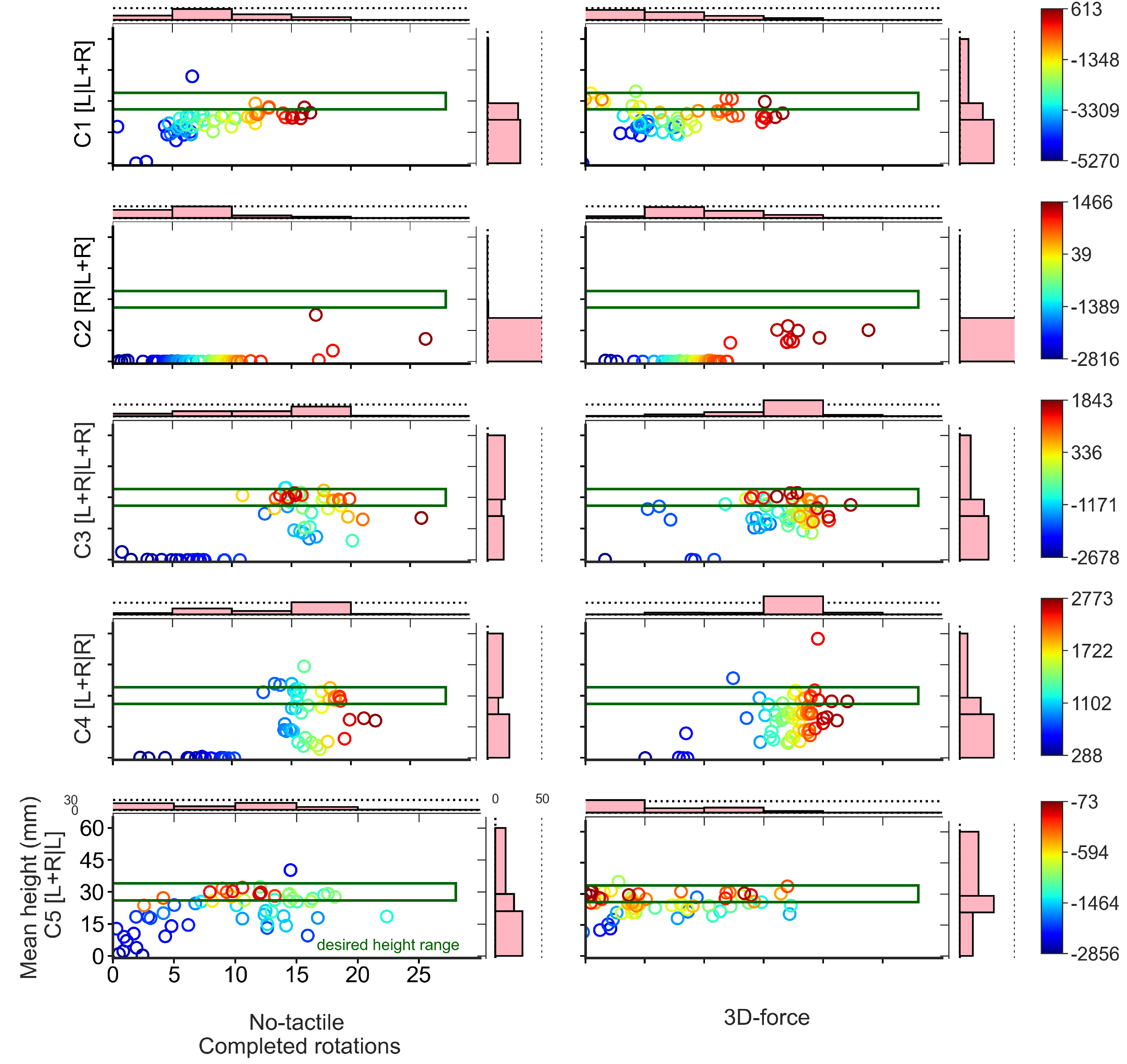}
    \caption[]  
    {\footnotesize \textbf{Evaluation of performance across all curricula and both tactile information options for Object 4: 5g, 30mm.}
     The joint distribution illustrates the performance during the final episode of 60 trial runs (showcasing the mean ball height (mm) versus the number of completed rotations). The color-coded cumulative reward for the last episode of each independent run (refer to equation 1) corresponds to different curricula. Note that the desired manipulation performance is represented by those points inside the green box defining the desired ball height (30 ± 4 mm).
    }
\label{fig: O4_jointplots}
\end{figure*}

\begin{figure*}[ht]
  \centering
      \includegraphics[width=0.85\textwidth]{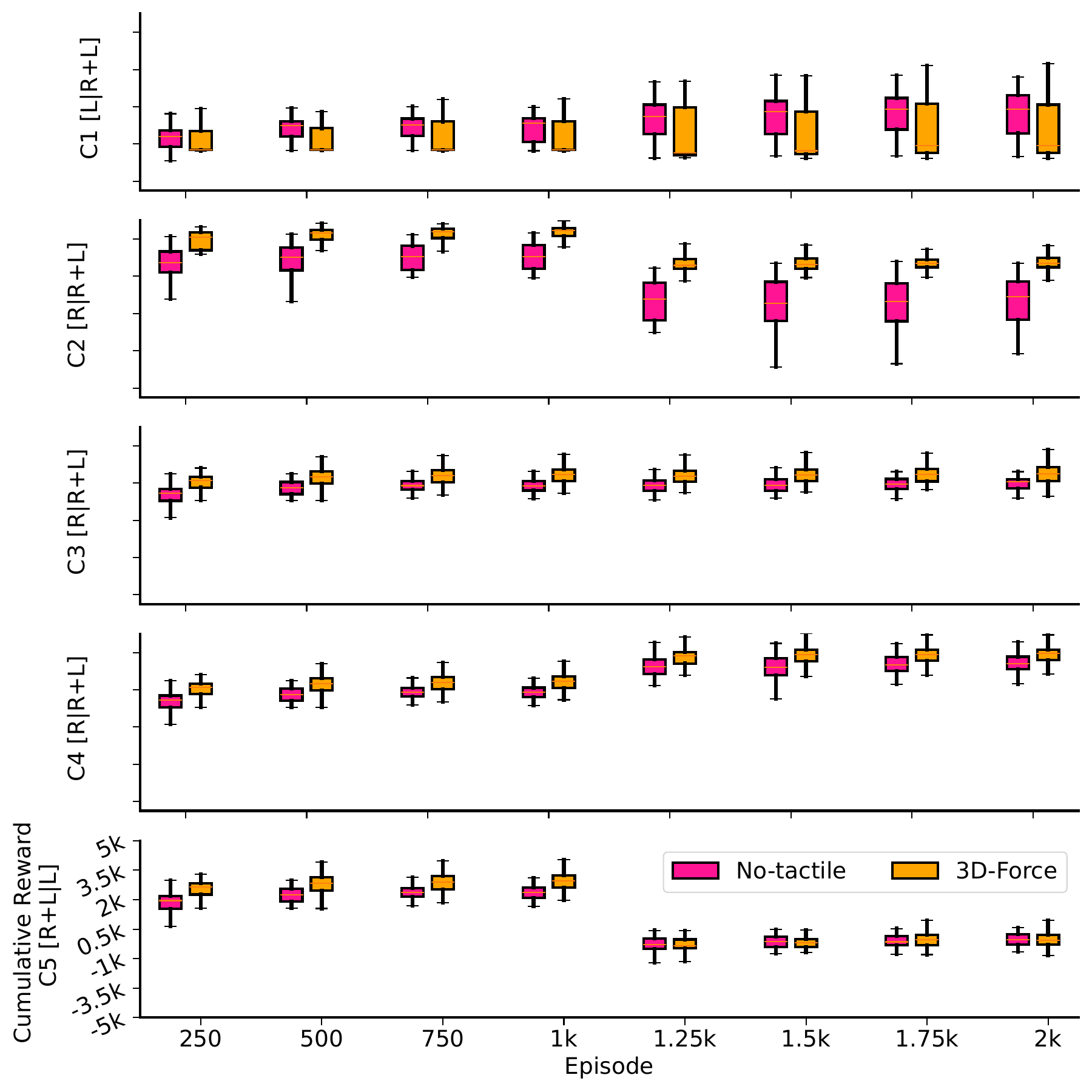}
     \caption[]
    {\footnotesize \textbf{Cumulative reward for each representative episode across all curricula and both tactile information options for Object 4: 5g, 30mm}. Boxplots, with median, across tactile information options for 60 MC runs at eight representative episodes, 250 episodes apart.
    }
 \label{fig:S4_boxplots}
\end{figure*} 

\newpage
\subsubsection*{Caption for Supplementary Video} 
\textbf{Video of the robotic hand interacting with the ball after undergoing learning} 
The ability of the robotic hand to manipulate was evaluated after 2,000 training episodes. 
Learning strategy influences manipulation performance.
How we started the learning had a direct effect on the same end goal. 
Here we showcase an example of learning in \Cthree and \Cfive.
Performance representative of each Curriculum while using 3D-force tactile information are shown in the video. 
Interesting insights were gleaned from each approach to autonomous learning of manipulation.

\end{document}